\documentclass{article}

\usepackage{microtype}
\usepackage{graphicx}
\usepackage{subfigure}
\usepackage{caption}
\usepackage{subcaption}
\usepackage{booktabs} %
\usepackage[normalem]{ulem}

\usepackage{hyperref}

\usepackage{authblk}

\usepackage[accepted]{styles/icml2024}

\usepackage{amsmath}
\usepackage{amssymb}
\usepackage{mathtools}
\usepackage{amsthm}
\usepackage{calc}

\usepackage{minitoc}

\usepackage{float}

\usepackage[capitalize,noabbrev]{cleveref}

\usepackage[capitalize]{cleveref}
\crefname{appendix}{App.}{Apps.}
\crefname{section}{Sec.}{Secs.}
\Crefname{section}{Sec.}{Sections}
\Crefname{table}{Table}{Tables}
\crefname{table}{Tab.}{Tabs.}
\Crefname{figure}{Fig.}{Figs.}
\Crefname{equation}{Eq.}{Eqs.}

\usepackage{amsmath,amsfonts,bm}

\def\eqref#1{equation~\ref{#1}}

\def\1{\bm{1}}

\def\rvb{{\mathbf{b}}}

\def\rvf{{\mathbf{f}}}

\def\rvw{{\mathbf{w}}}
\def\rvx{{\mathbf{x}}}

\def\rvz{{\mathbf{z}}}

\def\rmB{{\mathbf{B}}}

\def\rmD{{\mathbf{D}}}

\def\rmI{{\mathbf{I}}}

\DeclareMathAlphabet{\mathsfit}{\encodingdefault}{\sfdefault}{m}{sl}
\SetMathAlphabet{\mathsfit}{bold}{\encodingdefault}{\sfdefault}{bx}{n}

\def\gC{{\mathcal{C}}}

\def\gE{{\mathcal{E}}}

\def\gN{{\mathcal{N}}}

\def\sR{{\mathbb{R}}}

\newcommand{\E}{\mathbb{E}}

\newcommand{\R}{\mathbb{R}}

\newcommand{\KL}{D_{\mathrm{KL}}}

\DeclareMathOperator*{\argmin}{arg\,min}

\usepackage{url}
\usepackage{multirow}
\usepackage{multicol}
\usepackage{adjustbox}
\usepackage{float}

\theoremstyle{plain}
\newtheorem{theorem}{Theorem}[section]

\newtheorem{lemma}[theorem]{Lemma}

\theoremstyle{definition}

\newtheorem{assumption}[theorem]{Assumption}
\theoremstyle{remark}

\usepackage[textsize=tiny]{todonotes}

\newif\ifShowComments

\newif\ifHighlightRebuttal
\newcommand{\rnew}[1]{
    \ifHighlightRebuttal{\textcolor{red}{#1}}
    \else{#1}
    \fi
}

\icmltitlerunning{Align Your Steps: Optimizing Sampling Schedules in Diffusion Models}

\begin{document}
\twocolumn[{
\icmltitle{\textit{Align Your Steps}: Optimizing Sampling Schedules in Diffusion Models}

\vspace{-3mm}

\icmlsetsymbol{equal}{*}

\begin{icmlauthorlist}
\icmlauthor{Amirmojtaba Sabour$^*$}{nvidia,toronto,vector}
\icmlauthor{Sanja Fidler}{nvidia,toronto,vector}
\icmlauthor{Karsten Kreis}{nvidia}
\end{icmlauthorlist}

\icmlaffiliation{toronto}{Department of computer science, University of Toronto, Toronto, Ontario}
\icmlaffiliation{nvidia}{NVIDIA, Toronto, Canada}
\icmlaffiliation{vector}{Vector Institute, Toronto, Canada}

\icmlcorrespondingauthor{Amirmojtaba Sabour}{amsabour@cs.toronto.edu}

\icmlkeywords{Machine Learning, ICML, Generative models, Diffusion models, Efficient sampling}

\begin{center}
    \small 
    $^1$ NVIDIA \hspace{5mm}
    $^2$ University of Toronto \hspace{5mm}
    $^3$ Vector Institute
\end{center}

\begin{center}
\small{
\textit{Project Page: }
\href{https://research.nvidia.com/labs/toronto-ai/AlignYourSteps/}{
\textcolor{RubineRed}{https://research.nvidia.com/labs/toronto-ai/AlignYourSteps/}}
}
\end{center}

\vskip 0.3in
\begin{center}
  \includegraphics[width=\textwidth]{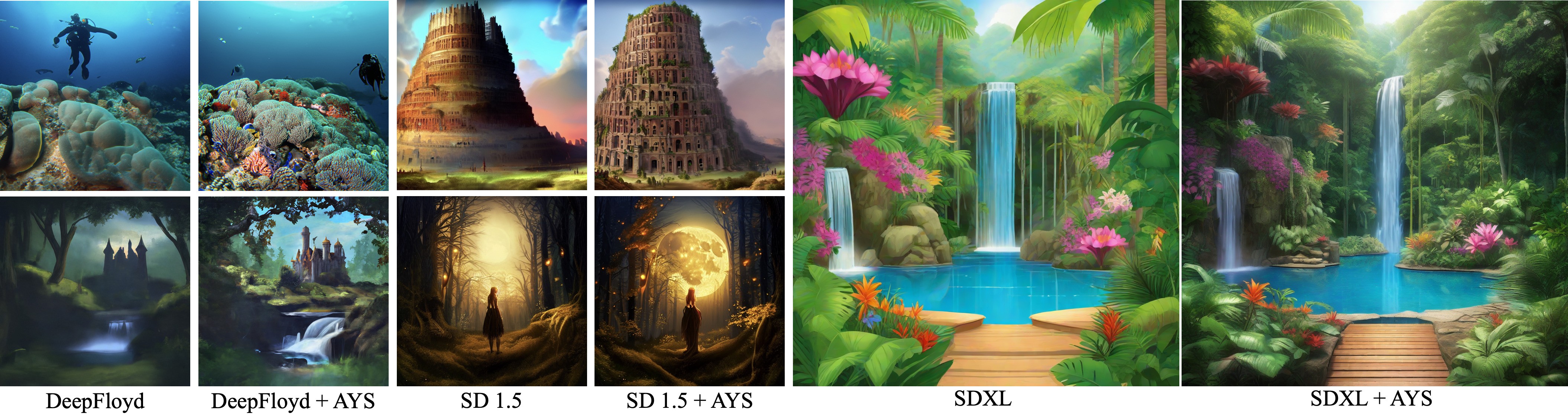}
  \vspace{-8mm}
  \captionof{figure}{
    We introduce \textit{\textbf{Align Your Steps} (AYS)}, a novel general framework for optimizing sampling schedules in diffusion models that significantly boosts the quality of outputs, especially when performing synthesis in few steps. Notice the improved details with AYS.
  }
  \label{fig:teaser}
\end{center}
\vskip 0.3in
}]

\printInternshipFootnote{}

\doparttoc %
\faketableofcontents %
\part{} %

\vspace{
-10mm
}
\begin{abstract}
\vspace{
    5mm
}
Diffusion models (DMs) have established themselves as the state-of-the-art generative modeling approach in the visual domain and beyond. A crucial drawback of DMs is their slow sampling speed, relying on many sequential function evaluations through large neural networks. Sampling from DMs can be seen as solving a differential equation through a discretized set of noise levels known as the sampling schedule. 
While past works primarily focused on deriving efficient solvers, little attention has been given to finding optimal sampling schedules, and the entire literature relies on hand-crafted heuristics. In this work, for the first time, we propose a general and principled approach to optimizing the sampling schedules of DMs for high-quality outputs, called \textit{Align Your Steps}. We leverage methods from stochastic calculus and find optimal schedules specific to different solvers, trained DMs and datasets. We evaluate our novel approach on several image, video as well as 2D toy data synthesis benchmarks, using a variety of different samplers, and observe that our optimized schedules outperform previous hand-crafted schedules in almost all experiments. Our method demonstrates the untapped potential of sampling schedule optimization, especially in the few-step synthesis regime. 

\end{abstract}

\section{Introduction}
Diffusion models (DMs) have proven themselves to be extremely reliable probabilistic generative models that can produce high-quality data. They have been successfully applied to applications such as image synthesis \citep{dhariwal2021diffusion,Ho2020DenoisingDP,scoresde,Rombach2021HighResolutionIS,Saharia2022PhotorealisticTD,Ramesh2022HierarchicalTI}, image super-resolution \citep{saharia2021image}, image-to-image translation \citep{Saharia2021PaletteID}, image editing \citep{brooks2022instructpix2pix}, inpainting \citep{Lugmayr2022RePaintIU},  video synthesis \citep{ho2022video,blattmann2023videoldm}, text-to-3d generation \citep{poole2022dreamfusion,lin2023magic3d}, and even planning \citep{janner2022diffuser}. 
However, sampling DMs requires multiple sequential forward passes through a large neural network, limiting their real-time applicability.

As a result, extensive research effort has gone into designing fast and efficient samplers of these models, broadly categorized into training-based and training-free methods. 
Training-based approaches, such as distillation, can significantly accelerate the sampling process but often require significant compute power, comparable to training the model itself, and face a trade-off between speed, diversity, and fidelity \citep{salimans2022progressive,song2023consistency,sdxlturbo,Luo2023LatentCM,Yin2023OnestepDW}, lagging behind standard DMs in terms of output quality, especially in large models. Although promising, these methods have not yet found wide-spread adoption by practitioners. 
On the other hand, since sampling from DMs corresponds to solving a generative Stochastic or Ordinary Differential Equation (SDE/ODE) in reverse time \citep{scoresde}, training-free methods usually seek to derive more efficient SDE/ODE solvers, making them more broadly applicable to different models with relative ease \citep{Lu2022DPMSolverAF, Lu2022DPMSolverFS, ddim, ersde,Xu2023RestartSF, edm}.\looseness=-1

\begin{figure}[t!]
    \includegraphics[width=1.0\columnwidth]{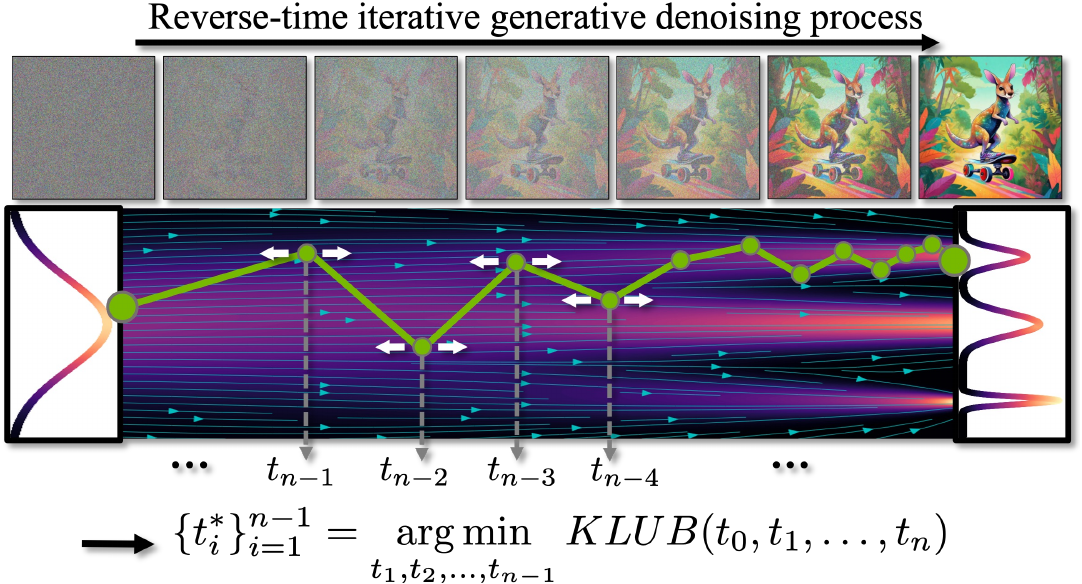}
  \vspace{-4mm}
    \caption{\small {\textbf{Align Your Steps.}} We minimize an upper bound on the Kullback-Leibler divergence ($\mathrm{KLUB}$) between the true and linearized generative SDEs to find optimal DM sampling schedules.}
    \label{fig:pipeline}
  \vspace{-3mm}
\end{figure}

Solving SDE/ODEs within the interval $[t_{min}, t_{max}]$ works by discretizing it into $n$ smaller sub-intervals $t_{min} = t_0 < t_1 < \dots < t_{n} = t_{max}$, and numerically solving the differential equation between consecutive $t_i$ values.  
This discretization has been given many names in the literature, \textit{e.g.} step size schedule, denoising schedule, timestep schedule, etc.\footnote{This is different from the noising schedule which specifies the amount of noise injection and scaling in the forward process. Please refer to \Cref{sec:background} for details.} We will be referring to it as the \textit{sampling schedule}. 
Changing the sampling schedule can significantly change the quality of the outputs \citep{edm}; however, most prior works simply adopt one of a handful of heuristic schedules, such as simple polynomials and cosine functions. Although significant effort has gone into developing faster solvers, little research has been conducted to optimize the sampling schedule. 
We attempt to fill this gap by introducing a principled approach for optimizing the schedule in a dataset-specific manner, resulting in improved outputs given the same compute budget. 
We'll be focusing on stochastic SDE solvers. These solvers excel in sampling from diffusion models due to their built-in error-correction, allowing them to outperform ODE solvers.

In a toy example using a Gaussian data distribution (\Cref{sec:gaussian_optimal_schedule}), we demonstrate the reliance of the optimal sampling schedule on the dataset characteristics and find that the optimal schedule significantly differs from heuristic sampling schedules used across the literature.
With this as motivation, we propose \textit{Align Your Steps} (AYS), a principled and general framework for optimizing the sampling schedule specific to any choice of dataset, model, and stochastic SDE solver. The framework is based on the observation that all stochastic SDE solvers can be reinterpreted as exactly solving an approximated linearized SDE on short intervals. This allows us to minimize the mismatch between solving the approximated linear SDE and the true generative SDE using techniques from stochastic calculus by framing it as an optimization problem over the sampling schedule (\Cref{fig:pipeline}). Although the framework assumes the use of stochastic SDE solvers, we empirically find that the optimized schedules generalize to several popular ODE solvers as well. The proposed framework is general and applicable to all DMs regardless of the data modality, and it is the first general schedule optimization framework that leads to improved output quality.

We empirically evaluate our method by optimizing the schedule for various datasets and models. These include 2D toy data, standard image datasets such as CIFAR10~\citep{cifar10}, FFHQ~\citep{karras2019style}, and ImageNet~\citep{deng2009imagenet}, large scale text-to-image models widely used by practitioners such as Stable Diffusion \citep{Rombach2021HighResolutionIS} and SDXL~\citep{Podell2023SDXLIL}, as well as the recent video DM Stable Video Diffusion \citep{Blattmann2023StableVD}. 
Our results show the practical advantages of optimizing the sampling schedule, ranging from fewer outliers in 2D point generation, enhanced quality in image generation, and improved temporal stability in video generation (\Cref{fig:teaser}).

\textbf{Contributions.}
\textit{(i)} We analytically establish the dependency of the optimal sampling schedule on the ground truth data distribution. 
\textit{(ii)} We introduce \textit{Align Your Steps}, a principled and general framework for optimizing the sampling schedule specific to any dataset, model and stochastic solver. 
\textit{(iii)} We improve upon previous heuristic sampling schedules for many popular stochastic and deterministic solvers, especially in the low NFE regime. 
\textit{(iv)} We provide the optimized schedules for several commonly used models in the appendix to allow for easy plug-and-play use by the research community.

\section{Background} \label{sec:background}

DMs are probabilistic generative models that inject noise into the data with a forward diffusion process and generate samples by learning and simulating a time-reversed backward diffusion process, initialized with a sample from a tractable distribution, \textit{e.g.} Gaussian noise. %
We adopt the framework of \citet{edm}, denote the data distribution by $p_{\textrm{data}}(\rvx)$ where $\rvx \in \R^d$, and define $p(\rvx; \sigma)$ as the distribution obtained by adding i.i.d. Gaussian noise of standard deviation $\sigma$ to the data. 

\textbf{Forward process.}
Score-based diffusion models~\citep{scoresde} progressively transform the data $p_{\textrm{data}}(\rvx)$ towards a noise distribution through a forward noising process. 
This process is determined by a \textit{noising schedule}, consisting of two functions $s(t), \sigma(t)$ that define the scaling and noise level at time t. Specifically, $\rvx_t = s(t) \hat{\rvx}_t$ where $\hat{\rvx}_t \sim p(\rvx, \sigma(t))$. The distribution of $\rvx_t$ is denoted as $p'(\rvx, t)$.
Given the noising schedule, the forward noising process can be written in the form of the following SDE
\begin{equation} \label{eq:forward_sde}
    d\rvx_t = \frac{\dot{s}(t)}{s(t)} \rvx_t + s(t) \sqrt{2 \sigma(t) \dot{\sigma}(t)} d\rvw_t, 
\end{equation}
where $\rvw_t \in \R^d$ denotes a standard Wiener process.

\textbf{Backward process and sampling.}
The forward SDE in \Cref{eq:forward_sde} has an associated reverse-time diffusion process~\citep{scoresde} given by
\begin{equation} \label{eq:backward_sde}
\begin{split}
    d\rvx_t = & \left[ \frac{\dot{s}(t)}{s(t)} \rvx_t - 2 s(t)^2 \sigma(t) \dot{\sigma}(t) \nabla_x \log p\left(\frac{\rvx_t}{s(t)}, \sigma(t)\right) \right] dt \\ 
    & + s(t) \sqrt{2 \sigma(t) \dot{\sigma}(t)} d\Bar{\rvw}_t,
\end{split}
\end{equation}
where $\Bar{\rvw}_t$ denotes a standard Wiener process backwards in time. 
However, there exists an entire class of reverse-time SDEs with matching marginals as the backward SDE in \Cref{eq:backward_sde} \citep{Huang2021AVP, edm, ersde}. The most notable being the non-stochastic probability flow ODE, introduced by \cite{scoresde}:
\begin{equation}\label{eq:prob_flow_ode}
    d\rvx_t = \left[ \frac{\dot{s}(t)}{s(t)} \rvx_t - s(t)^2 \sigma(t) \dot{\sigma}(t) \nabla_x \log p\left(\frac{\rvx_t}{s(t)}, \sigma(t)\right) \right] dt.
\end{equation}

As stated previously, sampling from a diffusion model boils down to solving one of these SDE/ODEs backward in time starting from random noise. This is done by discretizing the interval $[t_{min}, t_{max}]$ into $n$ sub-intervals $t_{min} = t_0 < t_1 < \dots < t_n = t_{max}$, known as a sampling schedule, and solving the SDE/ODEs on this schedule.

\begin{figure}[t]
    \centering
    \includegraphics[width=0.9\linewidth]{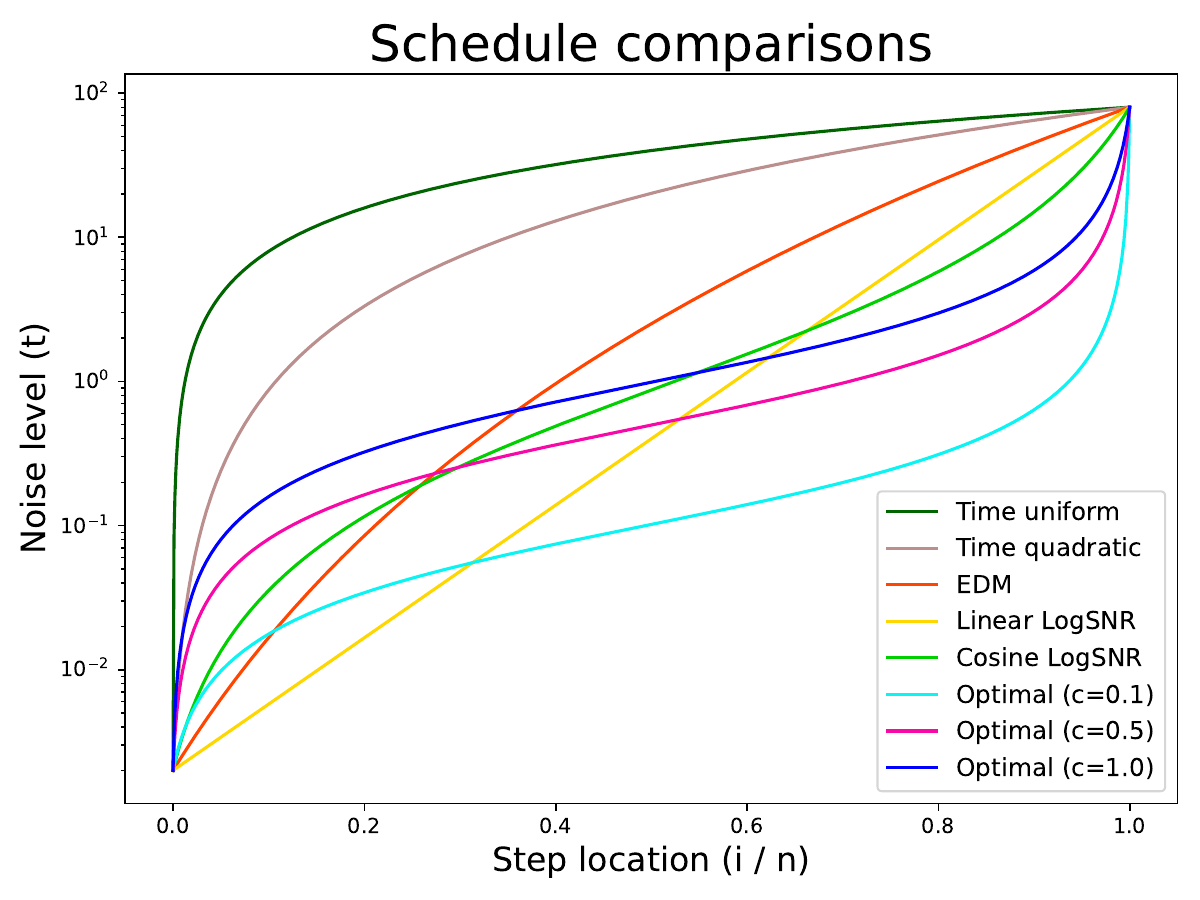}
    \caption{
        Comparing popular sampling schedules against the optimal schedules for Gaussian data.
    }
    \label{fig:compare_schedule_gaussian}
\end{figure}
\section{Optimizing Sampling Schedules}
Contrary to previous works, which have primarily focused on deriving efficient SDE/ODE solvers using heuristic schedules for sampling, we focus on fundamentally optimizing the sampling schedule given a specific choice of (dataset, model, stochastic solver) for a large class of SDE solvers.

In \Cref{sec:gaussian_optimal_schedule}, we first show how changing dataset characteristics causes the optimal sampling schedule to change.
Next, in \Cref{sec:KLUB}, we analyze the error introduced by discretizing the interval of the SDE into $n$ sub-intervals that define the sampling schedule, and formulate finding an optimal schedule as an optimization problem which can be solved iteratively.
\Cref{sec:importance_sampling} addresses implementation details.

\begin{figure*}[t]
    \centering
    \begin{minipage}{0.225\linewidth}
        \centering
        \includegraphics[width=\linewidth]{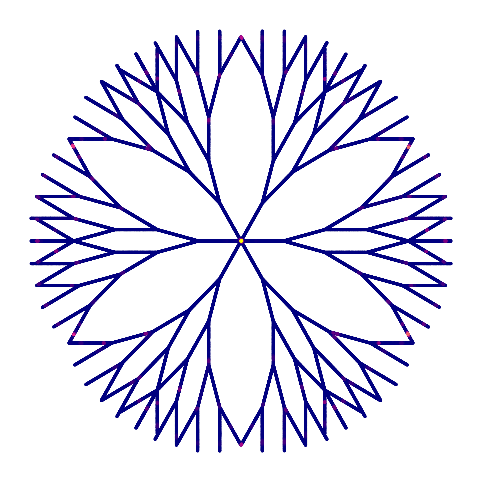} 
        \put(-100,10){(a)}
    \end{minipage}%
    \begin{minipage}{0.225\linewidth}
        \centering
        \includegraphics[width=\linewidth]{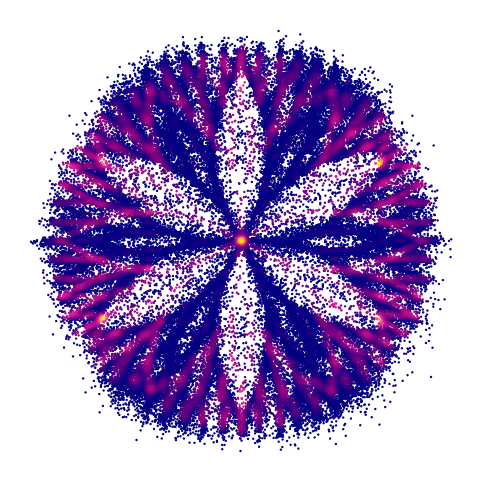} 
        \put(-100,10){(b)}
    \end{minipage}%
    \begin{minipage}{0.225\linewidth}
        \centering
        \includegraphics[width=\linewidth]{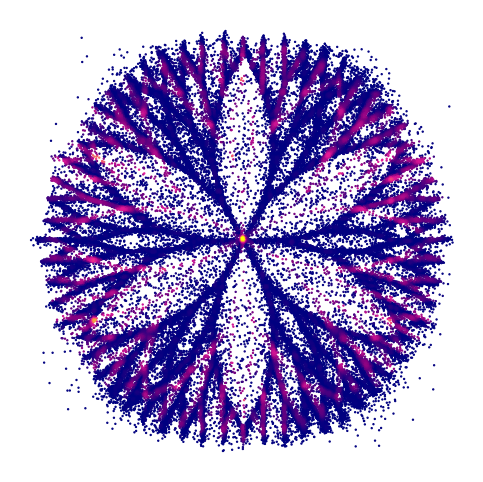} 
        \put(-100,10){(c)}
    \end{minipage}%
    \begin{minipage}{0.225\linewidth}
        \centering
        \includegraphics[width=\linewidth]{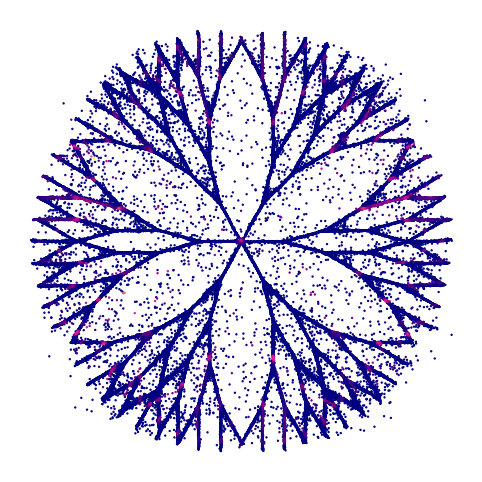} 
        \put(-100,10){(d)}
    \end{minipage}%
    \caption{
        \textbf{Modeling a 2D toy distribution}: (a) Ground truth samples; (b), (c), and (d) are samples generated using 8 steps of SDE-DPM-Solver++(2M) with EDM, LogSNR, and AYS schedules, respectively. Each image consists of 100,000 sampled points. The colors denote the local density of the samples where warmer colors correspond to higher density regions. See \Cref{appendix:extra_2d_experiments} for details.
    }    
    \label{fig:toy_samples}
\end{figure*}
\subsection{The Need for Optimized Schedules} \label{sec:gaussian_optimal_schedule}
Although the sampling schedule used for solving SDE/ODEs is a powerful hyperparameter at our disposal, little research effort has gone into optimizing it.
Especially in the relevant few-step synthesis regime, discretization errors can become significant~\citep{numericalmethods} and having an optimal sampling schedule can make a considerable impact.

As a motivating example, we analyze a simple case where an optimal sampling schedule can be derived analytically.
Consider the case where the initial distribution is an isotropic Gaussian with a standard deviation of $c$, \textit{i.e.} $p_{data}(\rvx) \sim \gN(\mathbf{0}, c^2 \rmI)$.  
We'll assume
$s(t) = 1, \sigma(t) = t$~\cite{edm}.
Forward SDE and Probability Flow ODE then are
\begin{equation}\label{eq:edm_forward}
\begin{cases}
    \text{Forward SDE:} & d\rvx_t = \sqrt{2t} \ d\rvw_t, \\
    \text{Reverse ODE:} & d\rvx_t = -t\nabla_x\log p(\rvx_t, t) dt. \\
\end{cases}
\end{equation}

In this setting, assuming use of the forward Euler method, also known as DDIM~\cite{ddim}, to solve the reverse ODE, an optimal schedule can be derived analytically.
\begin{theorem}[Proof in \Cref{appendix:optimal_gaussian}] \label{theorem:optimal_gaussian}
     Let $p_{data}(\rvx) = \gN(\mathbf{0}, c^2 \mathbf{I})$. Sample $\rvx_{t_{max}}{\sim} p(\rvx, t_{max})$ and solve the probability flow ODE using $n$ forward euler steps along the schedule $t_{max} = t_n > t_{n-1} > \dots > t_{1} > t_{0} = t_{min}$ to obtain $\Bar{\rvx}_{t_{min}}$. The optimal schedule $t^*$ minimizing the KL-divergence between $p(\rvx, t_{min})$ and the distribution of $\Bar{\rvx}_{t_{min}}$ is given by
     \begin{gather*}
         \alpha_{min} \coloneqq \arctan (t_{min} /  c), \ \ \ \  \alpha_{max} \coloneqq \arctan ( t_{max} / c ) \\
         \Rightarrow t^*_i = c \tan \left(
         (1 - \frac{i}{n}) \times \alpha_{min} + (\frac{i}{n}) \times \alpha_{max}
         \right). 
     \end{gather*}
\end{theorem}

In this theorem, the distribution $p(\rvx, t_{min})$ is the output distribution of exactly solving the probability flow ODE from $t_{max}$ to $t_{min}$. Therefore, the theorem states that the optimal schedule $t^*$, that has the minimum mismatch between its outputs and the outputs of exactly solving the ODE, has the interesting property that $\arctan(t^* / c)$ is a linear function. 

In \Cref{fig:compare_schedule_gaussian}, we compare several popular sampling schedules used in practice against these optimal schedules when $t_{min} = 0.002, t_{max}=80.0$ for various initial std. devs. $c$. The featured schedules include EDM~\citep{edm}, Linear LogSNR~\citep{Lu2022DPMSolverAF,Lu2022DPMSolverFS}, Cosine LogSNR \citep{Hoogeboom2023simpleDE, Nichol2021ImprovedDD}, linear time \citep{ddim}, and quadratic time~\cite{ddim}. 
This plot shows how changing the dataset (through changing the data distribution's std. dev. $c$) can have a significant impact on the optimal sampling schedule. Judging by how dissimilar the hand-crafted schedules appear compared to the optimal Gaussian ones, it is reasonable to believe that optimizing the schedules for each dataset could lead to significant performance gains. 
Note that in practice, it is common to normalize the input data to ensure unit variance. Yet, even only comparing the optimal schedule when $c=1$ to the others, there remains a big difference between them.  
We show the distribution of outputs for different samplers in \Cref{appendix:gaussian_data_extras}.\looseness=-1

\subsection{Analyzing the Discretization Errors} \label{sec:KLUB}
Since the sampling schedule defines how the reverse-time generative SDE will be discretized, optimizing the schedule corresponds directly to minimizing the discretization error of solving the SDE/ODE.
One method for analyzing such discretization errors in diffusions (and SDEs in general) is to use Girsanov's theorem~\citep{oksendal}. 
A simplified version of Girsanov's theorem is the following: 
\begin{theorem}[KL-divergence Upper bound ($\mathrm{KLUB}$), proof in \Cref{appendix:proving_klub}] \label{theorem:klub}
    Consider the following two SDEs:
    \begin{equation*}
        \begin{cases}
            \text{SDE 1}: & d\rvx_t = \rvf_1(\rvx_{0 \to t}, t)dt + g(t) d\rvw_t \\
            \text{SDE 2}: & d\rvx_t = \rvf_2(\rvx_{0 \to t}, t)dt + g(t) d\rvw_t \\
        \end{cases}
    \end{equation*}
    where $\rvx_{0 \to t}$ represents the entire path from the start ($t=0$) to the current time $t$ (this formulation is useful for multi-step methods that benefit from having access to the history). 
    Let $P_1$ and $P_2$ be the resulting probability distributions at time $T$ of the outputs of SDE 1 and SDE 2, respectively. 

    Under mild regularity constraints, we have:
    \begin{equation}
    \begin{gathered}
        \KL (P_1 \Vert P_2) \leq \mathrm{KLUB}(0, T) \coloneqq \\
        \frac{1}{2} \E_{P_1^{\textrm{paths}}} \left[  \int_{0}^{T} \frac{||\rvf_1(\rvx_{0 \to t}, t) - \rvf_2(\rvx_{0 \to t}, t)||^2}{g(t)^2} dt\right], 
    \end{gathered}
    \end{equation}
    
    where $P_1^{\textrm{paths}}$ refers to the distribution over path space $\rvx_{0 \to T} \in \gC ([0, T]; \sR^d)$ generated by running SDE 1. 
\end{theorem}

This theorem gives us an upper bound on the outputs' mismatch of two SDEs that share a diffusion term. 
In this work, our main goal is minimizing the mismatch between the outputs obtained by exactly solving the reverse-time generative SDE without discretization and the outputs of stochastic SDE solvers in practice, which use a finite sampling schedule. 
Most stochastic solvers work by decomposing the problem into multiple sub-intervals, within each of which the SDE is approximated by a linear SDE that has the same diffusion term. For these linear SDEs, exact numerical solutions exist which are used by the solvers. 
Therefore, for each stochastic SDE solver there exists a solver-specific linearized SDE, and the outputs of these solvers are the exact solutions of their respective linearized SDEs.
As a result, we can use the theorem above to derive a Kullback-Leibler divergence Upper Bound ($\mathrm{KLUB}$) between the outputs of practical stochastic solvers and the outputs of solving the reverse-time generative SDE without discretization. To clarify, solving the generative SDE without discretization is not possible in practice due to the nonlinear nature of the neural network. However, Girsanov's theorem offers us a tool to analyze the corresponding distribution regardless.

In the following, we will demonstrate deriving the $\mathrm{KLUB}$ for Stochastic-DDIM ($\eta = 1$) \citep{ddim}, and a similar procedure can be applied to other solvers with minimal adjustments.
We follow \citet{edm} and let $\rmD_\theta(x, \sigma)$ be the learnt denoiser function that takes in a noisy sample $\rvx$ with $\sigma$ noise and denoises the sample. %
Plugging in the relation $\nabla_x \log p_\theta \left( \rvx, \sigma \right) = (D_\theta(\rvx, \sigma) - \rvx) / \sigma^2$ into \Cref{eq:backward_sde} yields the following \textit{true learnt SDE}:
\begin{equation}\label{eq:stochastic_ddim_sde} 
\resizebox{\columnwidth}{!}{$
\begin{aligned}
    d\rvx_t &= \left[ \left( \frac{\dot{s}(t)}{s(t)} + \frac{2s(t)^2 \dot{\sigma}(t)}{\sigma(t)} \right) \rvx_t  
    - \frac{2s(t)^2 \dot{\sigma}(t)}{\sigma(t)} 
    \rmD_\theta\left({\color{Green} \frac{\rvx_t}{s(t)}, \sigma(t)}\right) 
    \right] dt \\
    &+ s(t) \sqrt{2 \sigma(t) \dot{\sigma}(t)} d\Bar{\rvw}_t 
\end{aligned}
$}
\end{equation}

\citet{Lu2022DPMSolverFS} have shown that Stochastic-DDIM is the exact solution of a 1st order approximation of the true learnt SDE. This means when solving the SDE in the sub-interval $[t_{i-1}, t_{i}]$, using the assumption $\rmD_\theta(\frac{\rvx_t}{s(t)}, \sigma(t)) \approx \rmD_\theta(\frac{\rvx_{t_i}}{s(t_i)}, \sigma(t_i))$, the \textit{discretized learnt SDE} of this solver is %
\begin{equation} \label{eq:stochastic_ddim_linearized_sde}
\resizebox{\columnwidth}{!}{$
\begin{aligned}
    d\rvx_t &= \left[ \left( \frac{\dot{s}(t)}{s(t)} + \frac{2s(t)^2 \dot{\sigma}(t)}{\sigma(t)} \right) \rvx_t - \frac{2s(t)^2 \dot{\sigma}(t)}{\sigma(t)} 
    \rmD_\theta\left({\color{Red} \frac{\rvx_{t_i}}{s(t_i)}, \sigma(t_i)}\right)
    \right] dt \\
    &+ s(t) \sqrt{2 \sigma(t) \dot{\sigma}(t)} d\Bar{\rvw}_t,
\end{aligned}
$}
\end{equation}
where the outputs of applying 1 step of Stochastic-DDIM from noise level $t_{i} \to t_{i-1}$ are the exact solution of this SDE. Note that this is a linear SDE since the denoiser does not rely on the current state $\rvx_t$, but only on the fixed $\rvx_{t_i}$ at the beginning of the interval, and its output can be treated as a constant vector inside the interval. Stitching together all these linear SDEs for the different sub-intervals gives us a \textit{general discretized learnt SDE} that corresponds to applying the solver using the entire sampling schedule.
\begin{figure}[t]
    \centering
    \includegraphics[width=0.95\linewidth]{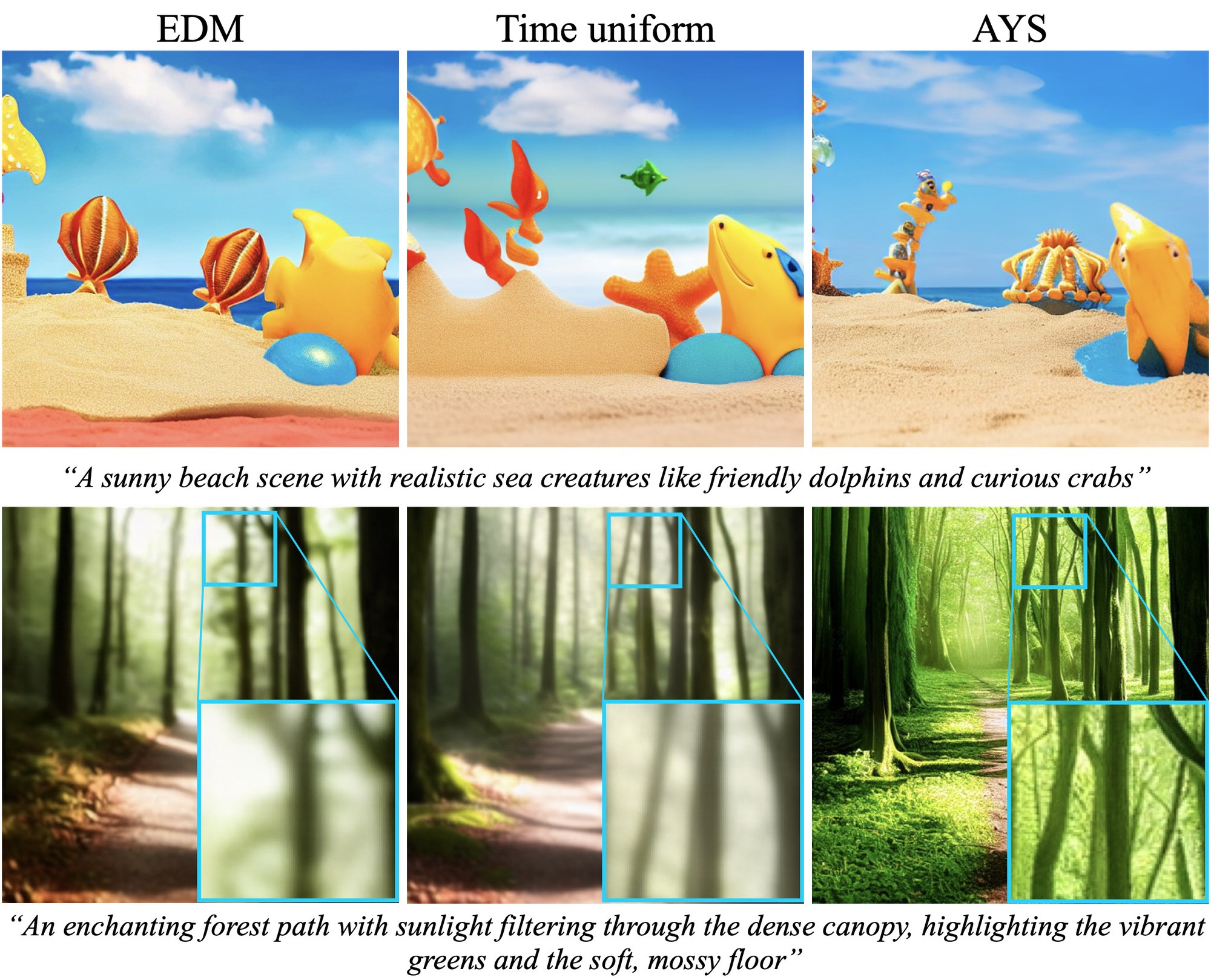}
    \caption{
    Side-by-side comparison of selected images generated with Stable Diffusion 1.5 with SDE-DPM-Solver++(2M) over 10 steps with different sampling schedules. 
    }
    \label{fig:sd_comparisons}
\end{figure}
\begin{figure}[t]
    \centering
    \includegraphics[width=0.9\linewidth]{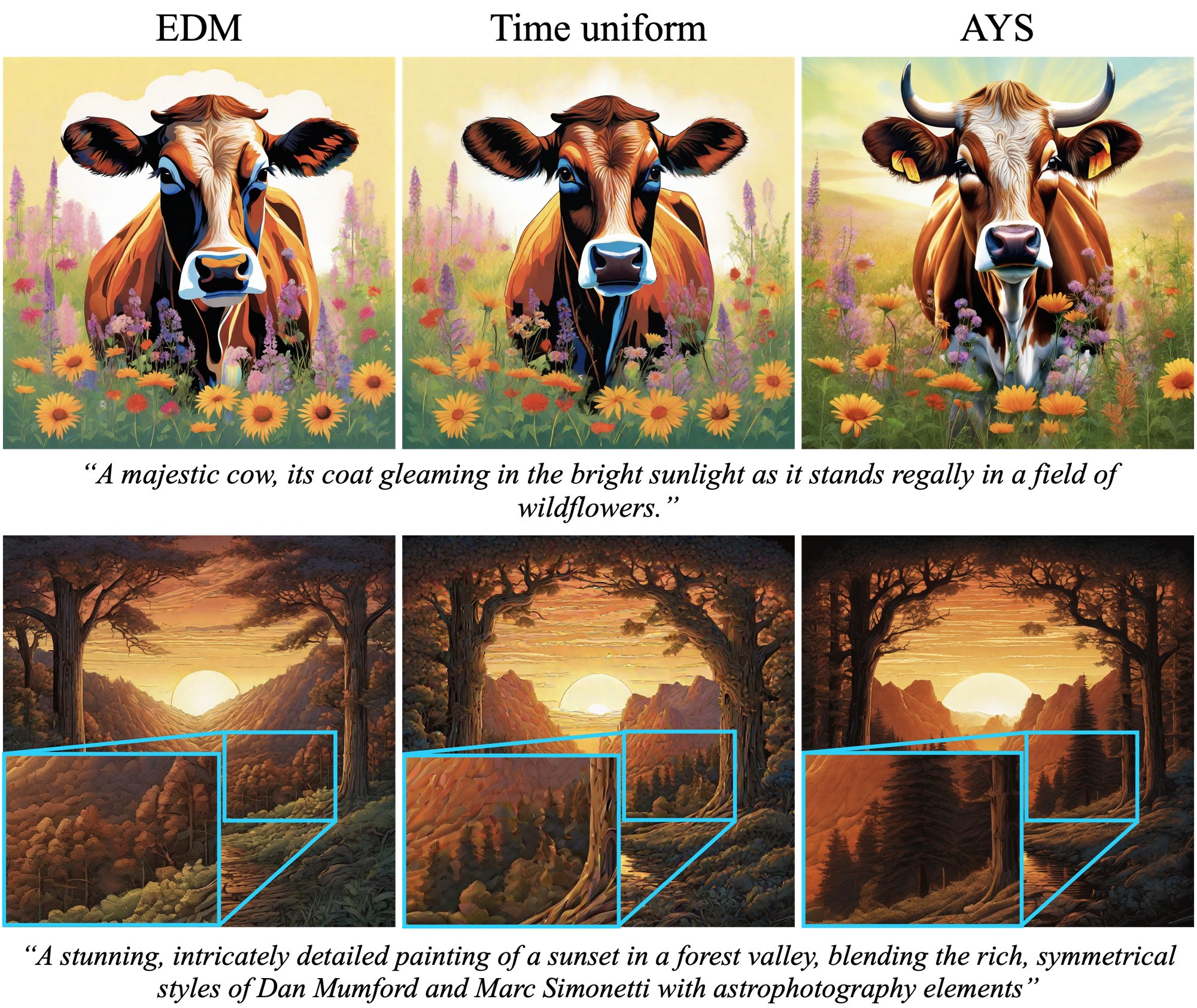}
    \caption{
    Side-by-side comparison of selected images generated with SDXL with 10 steps with different sampling schedules. The first and second rows use the SDE-DPM-Solver++(2M) and DPM-Solver++(2M) solvers respectively.
    }
    \label{fig:sdxl_comparisons}
\end{figure}

At this point, there are two SDEs that share the same diffusion term. The outputs of the true learnt SDE are samples obtained theoretically, given an unlimited compute budget, and outputs of the second general discretized SDE are samples obtained by running $n$ steps of Stochastic-DDIM along the finite sampling schedule in practice. 
The goal is to optimize the schedule in such a way as to ensure these two output distributions are as close as possible to each other, and for that we can use our $\mathrm{KLUB}$ formalism from above. 

To start, we consider a single sub-interval. Assuming both SDEs start from the forward diffusion process' distribution $p'(\rvx, t_{i})$ and are run from $t_{i} \to t_{i-1}$, 
we can apply \Cref{theorem:klub} backwards in time to obtain a $\mathrm{KLUB}$ between their output distributions. 
Letting the SDE in \Cref{eq:stochastic_ddim_sde} be $\text{SDE 1}$ and the SDE in \Cref{eq:stochastic_ddim_linearized_sde} be $\text{SDE 2}$ in the theorem, we obtain:
\begin{equation}
\begin{gathered}
    \KL (P_{t_{i} \to t_{i-1}}^{\textrm{true}} \Vert P_{t_{i} \to t_{i-1}}^{\textrm{disc}}) \leq \\
    \resizebox{\columnwidth}{!}{$
    2 \times \E_{P_{t_{i} \to t_{i-1}}^{\text{true paths}}} \int_{t_{i-1}}^{t_{i}} \frac{s(t)^2 \dot{\sigma}(t)}{\sigma(t)^3} \left\lVert \rmD_\theta\left({\color{Green}\frac{\rvx_t}{s(t)}, \sigma(t)}\right) -  \rmD_\theta\left({\color{Red} \frac{\rvx_{t_{i}}}{s(t_{i})}, \sigma(t_{i}})\right) \right\rVert^2 dt. $}
\end{gathered}
\end{equation}
Here $P_{t_{i} \to t_{i-1}}^{\textrm{true}}$ represents the distribution of running the true learnt SDE, $P_{t_{i} \to t_{i-1}}^{\textrm{disc}}$ denotes the distribution of running the discretized learnt SDE (that corresponds to Stochastic DDIM's 1-step outputs), and $P_{t_{i} \to t_{i-1}}^{\text{true paths}}$ is the distribution over path space of the true learnt SDE.

If we had a perfect score model, i.e. $\rmD_\theta(\rvx, \sigma) = E_{p_{\textrm{data}}(\rvx_0 | \rvx_\sigma)} [\rvx_0]$, then $P_{t_{i} \to t_{i-1}}^{\text{true paths}}$ would perfectly match the path distributions of the forward noising process, and $P_{t_{i} \to t_{i-1}}^{\textrm{true}} = p'(\rvx, t_{i-1})$, where $p'$ is the distribution of the forward noising process. 
We'll assume that $\rmD_\theta$ is sufficiently close to the true denoising function, and approximate it as such moving forward (for a more detailed error analysis, please refer to \Cref{appendix:error_analysis}). 
Applying this approximation to the equation above results in the following: 
\begin{equation}
\begin{gathered}
    \KL (P_{t_{i} \to t_{i-1}}^{\textrm{true}} \Vert P_{t_{i} \to t_{i-1}}^{\textrm{disc}}) \\
    \resizebox{\columnwidth}{!}{$
    \leq 2 \times \E_{P_{t_{i} \to t_{i-1}}^{\text{true paths}}} \int_{t_{i-1}}^{t_{i}} \frac{s(t)^2 \dot{\sigma}(t)}{\sigma(t)^3} \left\lVert \rmD_\theta\left({\color{Green} \frac{\rvx_t}{s(t)}, \sigma(t)}\right) - \rmD_\theta\left({\color{Red} \frac{\rvx_{t_{i}}}{s(t_{i})}, \sigma(t_{i})}\right) \right\rVert^2 dt$} \\
    \resizebox{\columnwidth}{!}{$
    \approx 2 \times \int_{t_{i-1}}^{t_{i}} \frac{s(t)^2 \dot{\sigma}(t)}{\sigma(t)^3} \E_{\substack{\rvx_t \sim p'(\rvx, t) \\ \rvx_{t_{i}} \sim p'(\rvx_{t_{i}} | \rvx_t)}} \left\lVert \rmD_\theta\left({\color{Green} \frac{\rvx_t}{s(t)}, \sigma(t)}\right) - \rmD_\theta\left({\color{Red} \frac{\rvx_{t_{i}}}{s(t_{i})}, \sigma(t_{i})}\right) \right\rVert^2 dt.$}
\end{gathered}
\end{equation}
This final value can be estimated using Monte Carlo integration and $(x_t, x_{t_{i}})$ can be drawn from the forward diffusion. %
\begin{figure*}[t]
    \centering
    \begin{minipage}{0.41\linewidth}
        \centering
        \includegraphics[width=\linewidth]{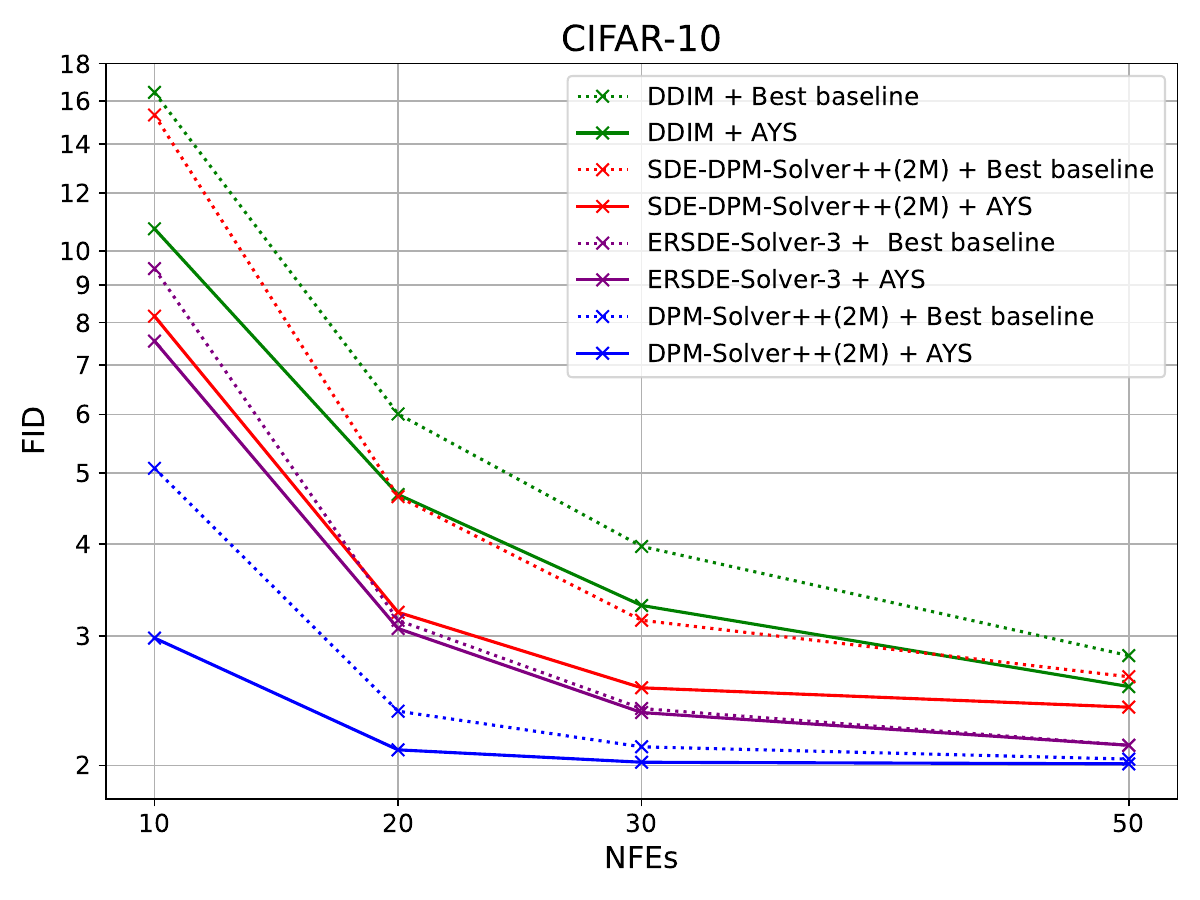} 
    \end{minipage}%
    \begin{minipage}{0.41\linewidth}
        \centering
        \includegraphics[width=\linewidth]{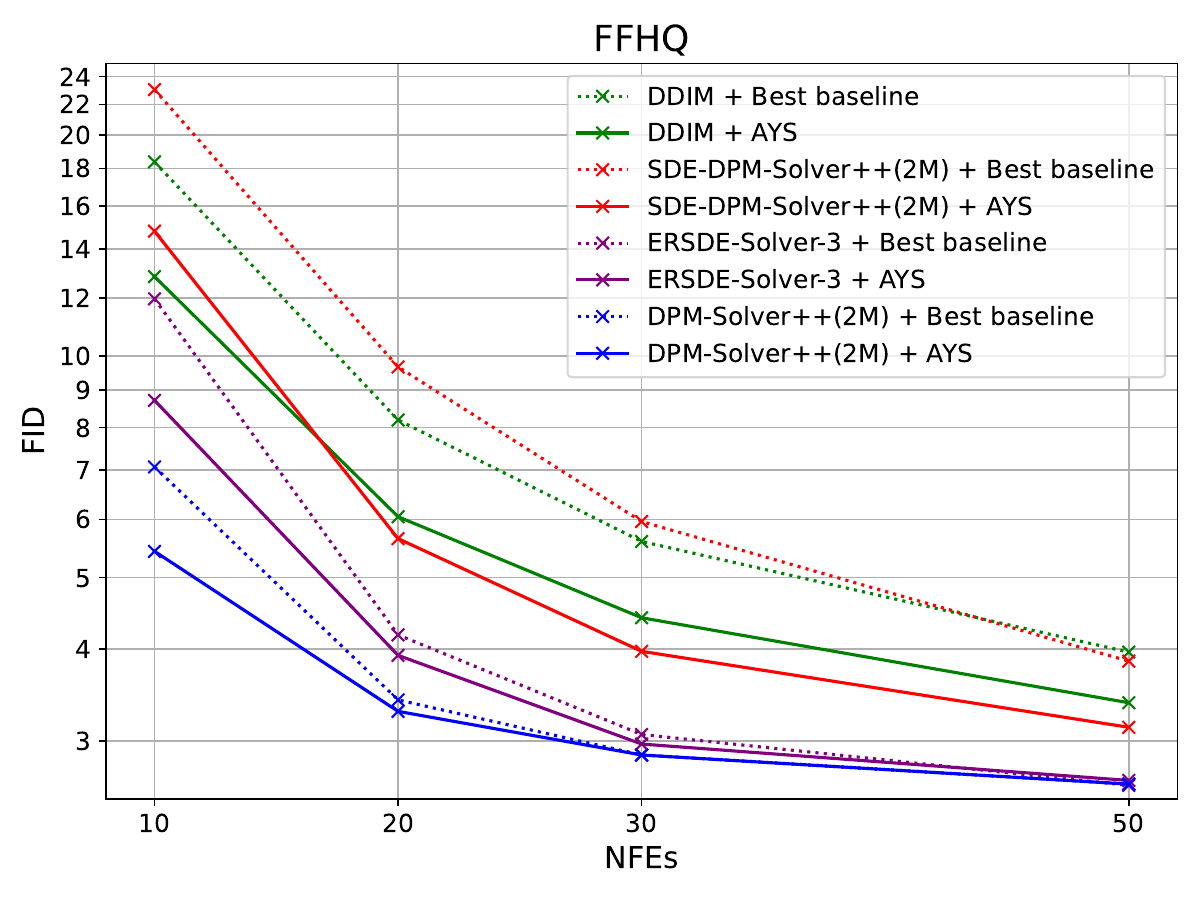} 
    \end{minipage}%
    \hfill
    \begin{minipage}{0.15\linewidth}
        \caption{
            FID curves for different solvers and schedules on CIFAR10 \textit{(left)} and FFHQ \textit{(right)}. See \Cref{table:cifar10_fids,table:ffhq_fids} in \Cref{appendix:edm_fid_tables} for more comprehensive results. 
        }  
        \label{fig:edm_fid_plots}
    \end{minipage}
\end{figure*}

This approach can be easily extended to the entire integration from $t_{max} \to t_{min}$. Assuming the sampling schedule is $t_{min} = t_0 < t_1 < \dots < t_{n} = t_{max}$, we apply the same technique on all sub-intervals and combine them to achieve a total $\mathrm{KLUB}$ between the outputs of running the true learnt SDE and the general discretized learnt SDE (which corresponds to Stochastic-DDIM with $n$-steps following the sampling schedule). The total $\mathrm{KLUB}$ then is %
\begin{equation*}
    \mathrm{KLUB}(t_0, t_1, \dots, t_{n}) \\
\end{equation*}
\begin{equation} \label{eq:klub_final}
\resizebox{\columnwidth}{!}{$
\begin{gathered}
    = \sum_{i=1}^{n} \int_{t_{i-1}}^{t_{i}} \frac{s(t)^2 \dot{\sigma}(t)}{\sigma(t)^3} \E_{\substack{\rvx_t \sim p'_t \\ \rvx_{t_{i}} \sim p'_{t_{i} | t}}} \left\lVert \rmD_\theta\left({\color{Green} \frac{\rvx_t}{s(t)}, \sigma(t)}\right) - \rmD_\theta\left({\color{Red} \frac{\rvx_{t_{i}}}{s(t_{i})}, \sigma(t_{i})}\right) \right\rVert^2 dt.
\end{gathered}
$}
\end{equation}
Note that each of the integrals only depends on the beginning and end of the intervals (due to the solver being first-order), allowing us to rewrite \Cref{eq:klub_final} as:
\begin{equation}
\begin{gathered}
    \mathrm{KLUB}(t_0, t_1, \dots, t_{n}) = \sum_{i=1}^{n} \mathrm{KLUB}(t_{i-1}, t_{i}) .
\end{gathered}
\end{equation}

Finally, we formulate the problem of finding an optimal sampling schedule as minimizing this KLUB value, resulting in the following optimization:
\begin{equation}
\begin{split}
    t^*_{1, \dots, n-1} &= \argmin_{t_1, t_2, \dots, t_{n-1}} \mathrm{KLUB}(t_0, t_1, \dots, t_{n}) \\
    &= \argmin_{t_1, t_2, \dots, t_{n-1}} \sum_{i=1}^{n} \mathrm{KLUB}(t_{i-1}, t_{i}) ,
\end{split}
\end{equation}
assuming $t_0 = t_{min}, t_n = t_{max}$ are fixed. This optimization is done iteratively by choosing one of the schedule indices $i \in \{1, \dots, n-1\}$, discretizing a neighbourhood around $t_i$ into several candidate points, computing the $\mathrm{KLUB}$ for each candidate, and setting $t_i$ to the candidate with the least value. Due to the decomposition, this process can be highly parallelized for non-neighbouring indices. 
A pseudocode is given in \Cref{appendix:importance_sampling}. 
We call this technique \textit{Align Your Steps} (AYS).

\subsection{Practical Considerations of KLUB Estimation} \label{sec:importance_sampling}
As discussed in the previous section, estimating the $\mathrm{KLUB}$ is the key to optimizing the sampling schedule. As such, an accurate estimator for the $\mathrm{KLUB}$ with low variance is required, and Importance Sampling with respect to time $t$ is used to achieve this.
Inspired by prior work~\citep{lsgm} we select the importance sampling distribution based on Gaussian data assumptions.
Specifically, we assume Gaussian data and analytically calculate all integration terms in \Cref{eq:klub_final}. Then we sample $t$ from a distribution whose probability density function (pdf) matches these calculated values, up to a constant factor. Empirically, we found that this approach significantly reduces the variance in our $\mathrm{KLUB}$ estimation and is effective across all datasets. 

Under the Gaussian data assumption, we have the following:
\begin{lemma}[Proof in \Cref{appendix:early_stopping_needed}]
    Let $p_{\textrm{data}}(\rvx) = \gN(\mathbf{0}, c^2 \rmI)$.  We assume $\rmD(\rvx, \sigma) = E_{p_{\textrm{data}}(\rvx_0 | \rvx_\sigma)}[\rvx_0]$ to be the ideal denoiser. Then for all $t < t_{i}$
    we have
    \begin{equation}
    \begin{gathered}
        \E_{\substack{\rvx_t \sim p'_t \\ \rvx_{t_i} \sim p'_{t_i | t}}} \left[ \left\lVert \rmD\left(\frac{\rvx_t}{s(t)}, \sigma(t)\right) - \rmD\left(\frac{\rvx_{t_i}}{s(t_i)}, \sigma(t_i)\right) \right\rVert^2 \right] \\
        = c^4 \left(\frac{1}{\sigma(t)^2 + c^2} - \frac{1}{\sigma(t_i)^2 + c^2}\right).
    \end{gathered}
    \end{equation}
\end{lemma}
And applying this lemma to \Cref{eq:klub_final} yields:
\begin{equation}
\resizebox{\columnwidth}{!}{$
    \textrm{KLUB} \propto \sum_{i=1}^{n} \int_{t_{i-1}}^{t_{i}} \frac{s(t)^2 \dot{\sigma}(t)}{\sigma(t)^3} \left(\frac{1}{\sigma(t)^2 + c^2} - \frac{1}{\sigma(t_i)^2 + c^2}\right) dt.
$}
\end{equation}

For simplicity, we will use $\sigma(t) = t, s(t) = 1$~\cite{edm} moving forward.
Considering an example case of $(t_{i-1}, t_{i}, t_{i+1}) = (0.1, 0.2, 0.5)$, the values from the integral above range 3 orders of magnitude $[0-1000]$, and if Monte Carlo integration were to be used naively in this case, the estimator would have a huge variance. 
To fix this, we perform importance sampling on $t$ according to the distribution $\pi(t)$ where 
\begin{equation}
    \pi(t) \propto \frac{1}{t^3} \left(\frac{1}{t^2 + c^2} - \frac{1}{t_i^2 + c^2}\right)
\end{equation}
for $c=0.5$. 
Given these $t$ samples, we average the reweighted integration terms $||\rmD_\theta(\rvx_t, t) - \rmD_\theta(\rvx_{t_i}, t_i)||^2 / (\frac{1}{t^2 + c^2} - \frac{1}{t_i^2 + c^2})$ which yields the final estimation of the $\mathrm{KLUB}$ (up to a constant).
This results in a much lower-variance estimator of the $\mathrm{KLUB}$.
A pseudocode and extra visualizations are given in \Cref{appendix:importance_sampling}.

\rnew{
In practice, the schedules are optimized in a hierarchical fashion. 
Specifically, we start with a 10-step schedule initialized using one of the heuristic schedules   $(t_0, t_1, \dots, t_{10})$. This is then iteratively optimized on all the 9 intermediate points $(t_1, t_2, \dots, t_9)$. At this initial stage, an early stopping mechanism is necessary to avoid over-optimizing, which is due to the optimization objective being an upper bound on the discretization error and not the error itself (see \Cref{appendix:early_stopping_needed} for a rigorous proof). 
After this process is finished, two rounds of subdivision and further fine-tuning are performed to obtain a 40-step schedule. 
Each time the schedule $(t_0, t_1, \dots, t_n)$ is subdivided to obtain a new schedule with twice the number of steps $(t'_0, t'_1, \dots, t'_{2n}$ where $t'_{2i} = t_{i}$ and $\log t'_{2i + 1} = 0.5 \times (\log t_i + \log t_{i+1})$. After a subdivision, the training process only focuses on further optimizing the newly added intermediate points (i.e. $t'_{2i+1}$) and keeps the other points frozen. This allows the general ``shape'' of the schedule to become fixed, removing the need for early stopping during these later stages. 
Finally, to obtain a schedule with a different number of steps than $[10, 20, 40]$, we view the 40-step schedule as a piece-wise log-linear function and interpolate it to match the number of desired number of steps. See \Cref{appendix:importance_sampling} for more details. 
}
All in all, the schedule optimization requires only a few iterations to converge (${<}300$).

\begin{figure*}[t]
    \centering
    \begin{minipage}{0.7\linewidth}
        \includegraphics[width=\linewidth]{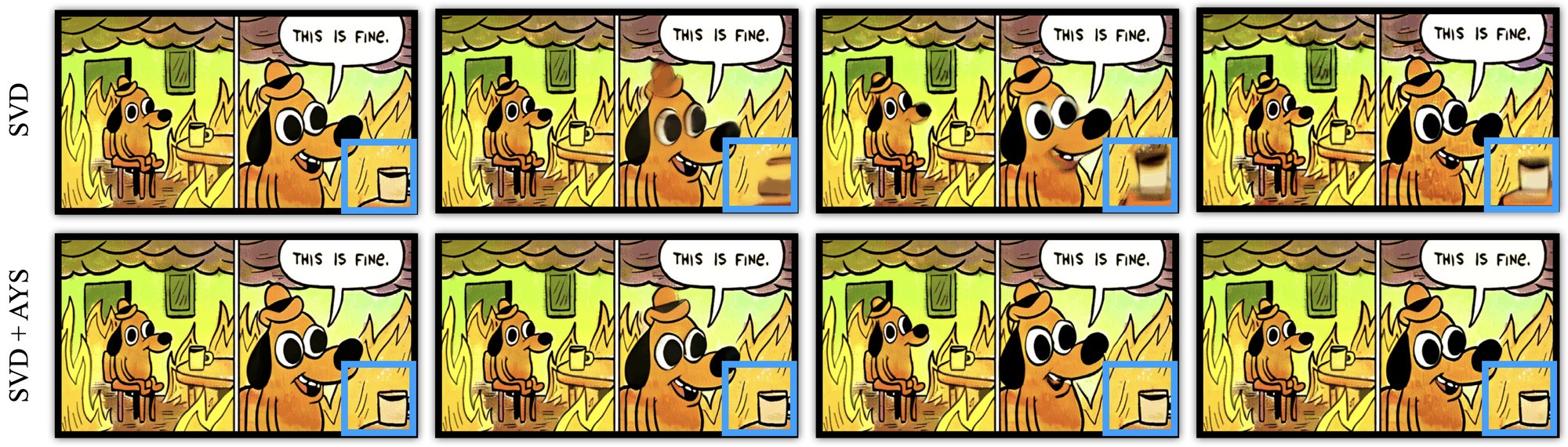}
    \end{minipage}\hfill
    \begin{minipage}{0.28\linewidth}
    \vspace{-2mm}
        \caption{
            Side-by-side comparisons for Stable Video Diffusion \citep{Blattmann2023StableVD}. We animate a meme (image-to-video). Using the optimized schedule results in a more stable video; note the temporal artifacts of the cup for the baseline. See supplementary material for full videos. 
        }
        \label{fig:svd_comparisons}
    \end{minipage}
\end{figure*}
\vspace{-1mm}

\section{Related Work}
We briefly review prior work on accelerating DM sampling.

Various \textbf{training-free methods} have been introduced to speed up DM synthesis, including efficient ODE~\citep{ddim,Lu2022DPMSolverAF, Lu2022DPMSolverFS,deis,dockhorn2022genie,liu2022pseudo, zheng2024dpm} and SDE solvers~\citep{JolicoeurMartineau2021GottaGF,Xu2023RestartSF}, as well as predictor-corrector methods \citep{scoresde,zhao2023unipc}. 
They are easy to integrate into existing models and we use several of these samplers in our experiments.

Moreover, \textbf{training-based methods} include neural operators \cite{Zheng2022FastSO}, truncated diffusion \cite{Zheng2022TruncatedDP, Lyu2022AcceleratingDM}, and distillation~\cite{salimans2022progressive,Meng2022OnDO,song2023consistency,Luo2023LatentCM,liu2023insta}, often employing adversarial objectives~\cite{xiao2022DDGAN,Xu2023UFOGenYF,Sauer2023AdversarialDD,Yin2023OnestepDW,kim2023consistency}. Although promising and almost reaching real-time sampling speeds, these methods often face trade-offs between inference speed, sample diversity, and output quality and require substantial compute for training. In practical applications, virtually all DMs rely on training-free samplers and solvers, which makes sampling schedule optimization a highly relevant task.

\citet{Watson2021LearningTE} introduced a dynamic programming method aimed at minimizing the DM's evidence lower bound (ELBO) to select the best $K$-step schedule from a larger $N$-step schedule. Although their optimized schedules improve log likelihoods, they do not yield improvements in image quality (as measured by FID scores). This is expected, as optimizing an exact ELBO is not favourable for image quality~\cite{Ho2020DenoisingDP}.
In follow-up work, \citet{Watson2022LearningFS} proposed differentiating through sample quality scores, specifically KID~\cite{binkowski2018demystifying}, to create an optimized sampler, including a trainable sampling schedule. This method showed improved FID scores compared to the baseline DDIM/DDPM samplers; however, it is limited to image-based diffusion models and lacks versatility for data types. 
Our method's comparison with this previous work can be seen in \Cref{appendix:google_comparisons}. 
In summary, we found their sampler to be outdated and it is unclear whether their optimized schedules are adaptable to different solvers. In contrast, our approach is derived in a principled manner, works on all data types, is compatible with a wide range of popular solvers, all while providing similar benefits. We also demonstrate our method on 2D data as well as video synthesis, which would not be possible with their technique. %
\rnew{
\citet{wang2023learning} explore the concept of asynchronous time inputs, where the time input provided to the denoiser differs from the actual noise level of the current latent, with these parameters being trainable and learned. This approach is orthogonal to ours, as it keeps a fixed ``sampling schedule'' while learning the ``denoiser inputs'', and integrating it with our optimized schedules could potentially improve the results even further.
\citet{xia2023towards} proposes using a schedule predictor, trained with reinforcement learning, that takes in the noisy latents and the current timestep as inputs, and predicts the optimal next step to denoise to. This results in a sampling schedule that adapts based on the sample being generated. However, the authors experiment exclusively with the first-order DDIM solver and it remains unclear if their learnt schedule predictor generalizes to more commonly used higher-order solvers.
}

\section{Experiments}
We demonstrate how optimizing the sampling schedule can significantly boost generation quality 
using the same number of forward evaluations (NFEs). 
We show how upsampling an optimized schedule with a small number of steps generalizes to higher NFE regimes as well as how using a schedule optimized on one solver's $\mathrm{KLUB}$ can generalize to other solvers. 
We compare outputs of various SDE/ODE solvers while using different schedules and show that optimized schedules lead to improvements almost across the board. 
Popular heuristic schedules listed in \Cref{appendix:popular_schedules}.

We evaluate our method on various datasets including 2D toy data, widely-used image datasets, and text-to-image and image-to-video models.
As sample quality metric, for CIFAR10~\cite{cifar10}, FFHQ~\cite{karras2019style}, and ImageNet~\cite{deng2009imagenet}, we use FID scores \citep{fid}. For the text-to-image and text-to-video models, we show the benefits of our method both qualitatively and quantitatively using human evaluation scores.

\begin{figure*}[t] 
\centering
\begin{minipage}[bt]{0.73\linewidth}
\centering
\captionof{table}{
    Sample fidelity (FID $\downarrow$, sFID $\downarrow$, Inception Score $\uparrow$) on the ImageNet $256 \times 256$ dataset. 
}
\label{table:imagenet_fids}
\resizebox{\linewidth}{!}{%
\begin{tabular}{@{}cllccccccccc@{}}
\toprule
                                         & \multirow{2}{*}{Sampling method}  & \multirow{2}{*}{Schedule}      & \multicolumn{3}{c}{NFE=10}  & \multicolumn{3}{c}{NFE=20} & \multicolumn{3}{c}{NFE=30}     \\ 
                                         \cmidrule(lr){4-6} \cmidrule(lr){7-9} \cmidrule(lr){10-12} 
                                         &  &                               & FID $\downarrow$ & sFID $\downarrow$ & IS $\uparrow$ 
                                                                            & FID $\downarrow$ & sFID $\downarrow$ & IS $\uparrow$
                                                                            & FID $\downarrow$ & sFID $\downarrow$ & IS $\uparrow$                                             \\
\midrule
\multirow{15}{2cm}{Stochastic Samplers}  & \multirow{3}{*}{Stochastic DDIM}        & EDM             & 66.71  & 126.92 & 25.04     & 17.42 & 49.89 & 152.74  & 9.85 & 26.15 & 242.81        \\ 
                                         &                                         & Time-uniform    & 24.48  & 67.96  & 112.53    & 9.32  & 22.65 & 256.27  & 8.41 & 13.67 & 299.44        \\ 
                                         &                                         & AYS             & \textbf{23.13}  & \textbf{64.37}  & \textbf{118.61}    & \textbf{8.96}  & \textbf{19.78} & \textbf{264.98}  & \textbf{8.29} & \textbf{11.65} & \textbf{304.37}        \\ 
                                         \cmidrule{2-12}
                                         & \multirow{3}{*}{SDE-DPM-Solver++ (2M)}  & EDM             & 8.48   & 21.83  & 214.49    & 7.05  & 8.17  & 307.41  & 7.55 & 6.58  & 325.78        \\ 
                                         &                                         & Time-uniform    & 8.47   & 13.36  & 243.09    & 7.63  & 11.02 & 282.77  & \textbf{7.14} & 8.59  & 305.57        \\ 
                                         &                                         & AYS             & \textbf{6.11}   & \textbf{8.48}   & \textbf{281.44}    & \textbf{6.79}  & \textbf{5.93}  & \textbf{322.92}  & 7.28 & \textbf{5.48}  & \textbf{330.01}        \\ 
                                         \cmidrule{2-12}
                                         & \multirow{3}{*}{ER-SDE-Solver 1}        & EDM             & 17.78  & 35.25  & 147.57    & 6.99  & 12.70 & 255.69  & 6.20 & 8.51  & 282.52        \\ 
                                         &                                         & Time-uniform    & 8.79   & 18.33  & 222.93    & 6.25  & 8.19  & 280.74  & 6.09 & 6.56  & 293.47        \\ 
                                         &                                         & AYS             & \textbf{8.36}   & \textbf{15.91}  & \textbf{266.44}    & \textbf{6.06}  & \textbf{7.28}  & \textbf{282.06}  & \textbf{5.87} & \textbf{5.97}  & \textbf{295.40}        \\
                                         \cmidrule{2-12}
                                         & \multirow{3}{*}{ER-SDE-Solver 2}        & EDM             & 7.36   & 14.19  & 231.46     & 5.58 & 6.33  & 290.80  & 5.85 & 5.69  & 299.12        \\ 
                                         &                                         & Time-uniform    & \textbf{5.28}   & \textbf{6.19}   & \textbf{277.57}     & 5.56 & 5.55  & 295.69  & 5.72 & 5.50  & 300.25        \\ 
                                         &                                         & AYS             & 5.38   & 6.24   & 275.35     & \textbf{5.45} & \textbf{5.19}  & \textbf{297.78}  & \textbf{5.71} & \textbf{5.16}  & \textbf{301.79}        \\ 
                                         \cmidrule{2-12}
                                         & \multirow{3}{*}{ER-SDE-Solver 3}        & EDM             & 6.94   & 13.01  & 237.70     & 5.58 & 6.13  & 292.75  & 5.87 & 5.61  & 299.33        \\ 
                                         &                                         & Time-uniform    & \textbf{5.13}   & \textbf{6.08}   & \textbf{277.65}     & 5.52 & 5.57  & 295.94  & \textbf{5.71} & 5.48  & 301.52        \\ 
                                         &                                         & AYS             & 5.28   & 6.10   & 275.80     & \textbf{5.47} & \textbf{5.17}  & \textbf{298.05}  & 5.73 & \textbf{5.14}  & \textbf{302.40}        \\ 
\midrule
\multirow{4}{2cm}{Deterministic Solvers} & \multirow{2}{*}{DDIM}                   & Time-uniform    & 7.57   & 14.53  & 224.50     & 5.39    & 7.08    & 273.33       & 5.23    & 5.87    & 283.27             \\
                                         &                                         & AYS             & \textbf{6.96}   & \textbf{12.21}  & \textbf{226.25}     & \textbf{5.09}  & \textbf{12.21} & \textbf{273.94}   & \textbf{4.99}    & \textbf{5.53}    & \textbf{283.37}             \\
                                         \cmidrule{2-12}
                                         & \multirow{2}{*}{DPM-Solver++ (2M)}      & LogSNR          & 4.82   & 6.83   & 252.71     & 4.81  & 5.41  & 287.20   & 4.98  & 5.22  & 288.81       \\
                                         &                                         & AYS             & \textbf{4.31}     & \textbf{6.64}     & \textbf{260.32}         & \textbf{4.70}    & \textbf{5.34}   & \textbf{284.17}       & \textbf{4.96}    & \textbf{5.15}    & \textbf{290.65}             \\
                                         
\bottomrule
\end{tabular}%
}
\end{minipage}\hfill
\begin{minipage}[bt]{0.24\linewidth}
    \centering
    \includegraphics[width=\textwidth]{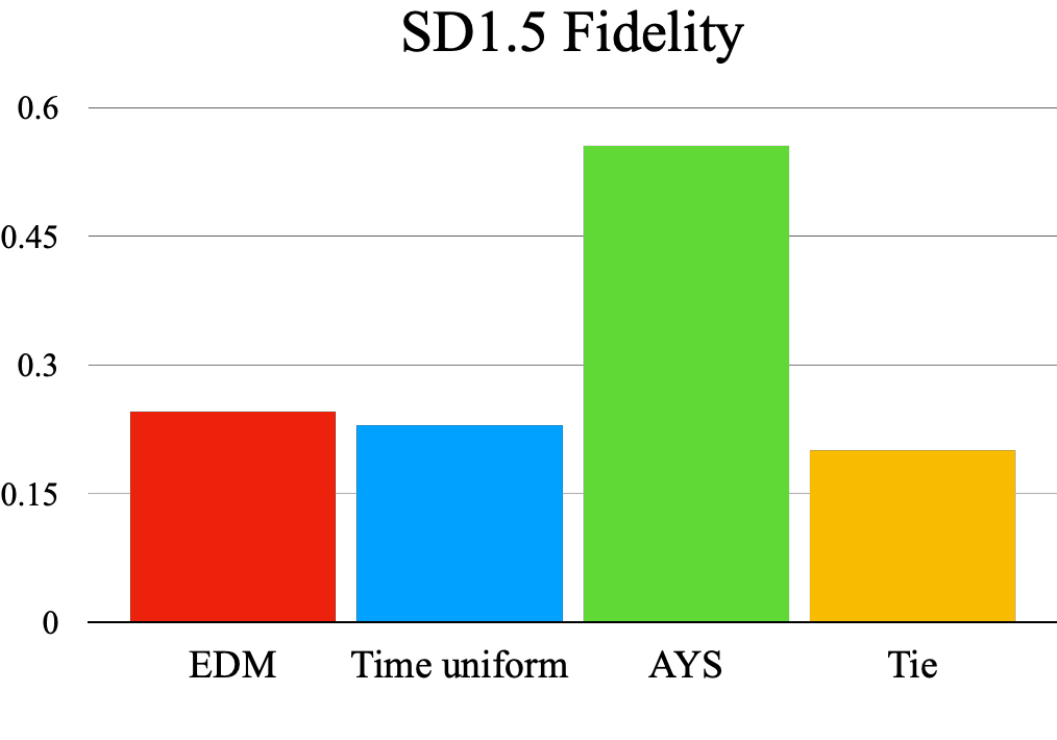}
    \includegraphics[width=\textwidth]{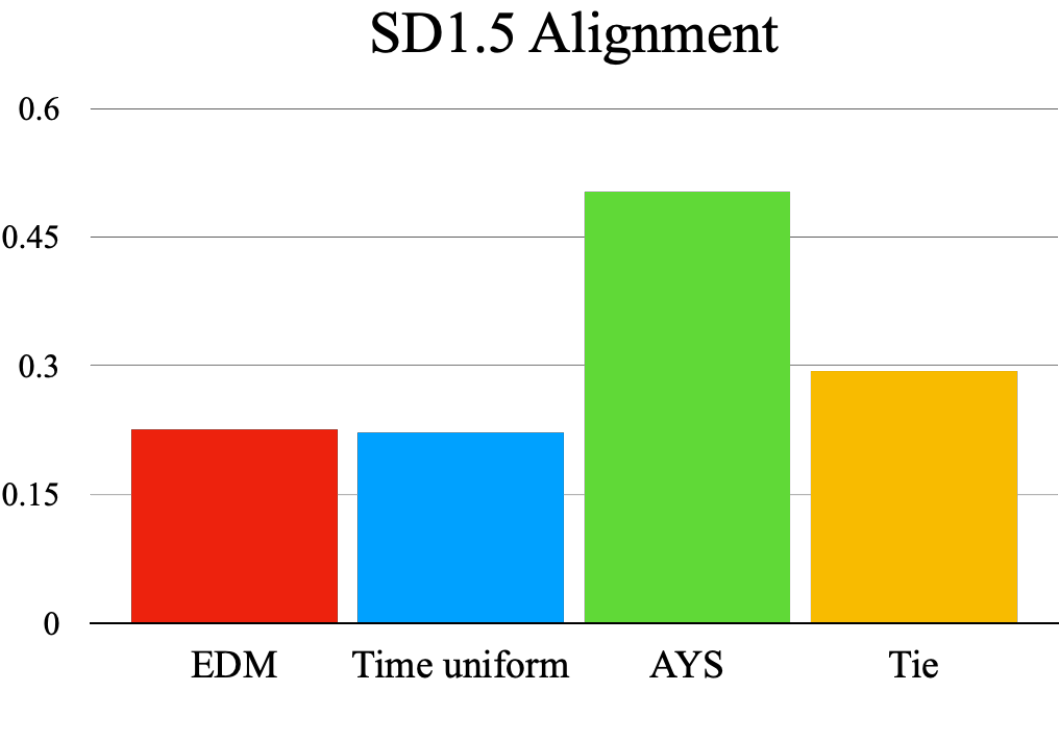}
    \caption{User study results on Stable Diffusion 1.5.}
    \label{fig:user_study}
\end{minipage}
\end{figure*}

\subsection{Toy Experiments}
In \Cref{fig:toy_samples}, we show the advantages of optimized sampling schedules using a 2D toy dataset. 
We used a continuous-time EDM-based DM to learn the score, which was used to optimize the schedule.
The samples generated with the optimized schedule more closely resemble the original distribution and have less outliers. 
Additional 2D results in \Cref{appendix:extra_2d_experiments}.\looseness=-1

\subsection{CIFAR10, FFHQ, ImageNet}
For CIFAR10 and FFHQ experiments, we use pretrained continuous-time DMs from \citet{edm}. %
For ImageNet, we use the pretrained latent DM from \cite{Rombach2021HighResolutionIS} with classifier free guidance with a scale of $2.0$. %

We use 3 different classes of stochastic solvers: Stochastic DDIM~\citep{ddim}, second-order SDE-DPM-Solver++ \citep{Lu2022DPMSolverFS}, and the recently proposed 1st, 2nd, and 3rd order ER-SDE-Solvers~\citep{ersde}. 
We also report FID scores for two popular deterministic solvers, namely  DDIM~\citep{ddim} and DPM-Solver++ (2M)~\citep{Lu2022DPMSolverFS}. 
For simplicity, no dynamic thresholding is used \citep{Saharia2022PhotorealisticTD}. %

In \Cref{fig:edm_fid_plots}, we compare FIDs of generated images using the AYS schedule versus the best baseline schedule across four different solvers, including two stochastic and two deterministic ones. 
The results clearly demonstrate the benefits of optimizing the schedule.
In some cases, \textit{e.g.} for SDE-DPM-Solver++(2M), images generated with an optimized 20 step schedule achieve FIDs comparable to those from a 30-step default schedule, achieving a 1.5x speedup. 
Additionally, the results indicate that as the number of steps increases, the impact of different schedules diminishes, which is due to the discretization error becoming small. 
For more comprehensive results, please see \Cref{table:cifar10_fids,table:ffhq_fids} in \Cref{appendix:edm_fid_tables}.  

In \Cref{table:imagenet_fids}, we compare the quality
of images generated using the EDM, time-uniform, and AYS schedules on ImageNet. 
While the FID values occasionally exhibit untypical behavior, such as deterioration with an increased number of steps, we suspect this is due to the absence of thresholding, potentially causing instabilities with higher-order solvers
for small NFE.
Nevertheless, in most instances, the optimized schedule outperforms the other two in all three metrics.

\subsection{Text-to-Image}
We also used our method to optimize sampling schedules for popular open-source text-to-image models, including Stable Diffusion 1.5 \citep{Rombach2021HighResolutionIS}, SDXL \citep{Podell2023SDXLIL}, and DeepFloyd-IF \citep{Deepfloyd}.
For models that rely on classifier-free guidance, each guidance value essentially creates a different score model, suggesting that the optimal schedule should be tailored to each specific value. However, our experiments show that schedules optimized with default guidance values are effective across a reasonable range of values. See \Cref{appendix:extra_txt2img} for FID vs. CLIP score pareto curves. 

The benefits of the optimized schedules are evident in \Cref{fig:sd_comparisons,fig:sdxl_comparisons}, which present side-by-side comparisons for SD 1.5 and SDXL, respectively. The results demonstrate that optimized schedules yield superior images in low NFE regimes, sometimes showing significant improvements. See \Cref{appendix:extra_txt2img} for additional side-by-side comparisons. 

\rnew{
To quantitatively evaluate the effectiveness of different schedules, we conducted a user study with 42 participants to assess \textit{image fidelity} and \textit{image-text alignment}. 
Each participant received a text prompt and three images generated with the EDM, time-uniform, and AYS schedules, respectively, using the same random seed. The SDE-DPM-Solver++(2M)~\citep{Lu2022DPMSolverFS} was used to generate the images with 10 steps. The order of the images was randomly permuted to avoid any biases. The participants then select the superior option according to image-quality and image-text alignment, or a choice for a three-way tie. 
The results, summarized in \Cref{fig:user_study}, reveal a clear preference for our optimized schedule with respect to both metrics.
}

\subsection{Video Generation Models}

With the growing interest in video synthesis and open-source video diffusion models becoming available, it is important to look at efficient samplers in this area. However, few efficient samplers have been evaluated in this context. 
To address this gap, we also study the effect of our method in this domain, using the recent Stable Video Diffusion (SVD) \citep{Blattmann2023StableVD}.
We compare videos generated using DDIM with the default EDM schedule against our optimized schedule in \Cref{fig:svd_comparisons}. 
We find that the optimized schedule helps improve temporal color consistency and addresses the issue of over-saturation in later video frames. 
\rnew{
We also conduct a user-study on the generated videos, similar to the image-generation case. However, due to the continuous nature of SVD, the EDM schedule is used by default and serves as the baseline, and we compare it against our optimized schedule. The default DDIM~\citep{ddim} was used with 10 steps to generate the videos due to the instability of higher-order solvers. Once again the results, summarized in \Cref{table:svd_user_study}, reveal a clear preference for our optimized sampling schedule.
} 
More details about these experiments in \Cref{appendix:video_model_details}.

\begin{table}[h]
\centering
\caption{Video generation user study results.}
\label{table:svd_user_study}
\scalebox{0.8}{
\begin{tabular}{@{}lcc@{}}
\toprule
        & EDM & AYS \\ 
\midrule
SVD \citep{Blattmann2023StableVD}        & 42\%             & 58\%      \\
\bottomrule
\end{tabular}}
\end{table}

\section{Conclusions and Future Work}

In summary, we present a novel framework for the optimization of sampling schedules in diffusion models aimed at enhancing the quality of generated samples in low NFE regimes. 
We successfully applied our method to several commonly used text-to-image and image-to-video models, and the schedules have been made publicly available\footnote{We also provide a colab notebook which shows how to use these schedules in practice on our \href{https://research.nvidia.com/labs/toronto-ai/AlignYourSteps/}{
\textcolor{RubineRed}{project page}}.}; see \Cref{appendix:popular_schedules} . 
\rnew{
Note that our framework is not strictly limited to diffusion models, and can also be integrated with recent, closely related generative techniques interpolating between data and noise, such as flow matching~\citep{lipman2022flow,esser2024scaling} and stochastic interpolants~\cite{albergo2023stochastic,ma2024sit}. When considering generative modeling with a Gaussian noise prior, these methods correspond to re-formulations of the same underlying generation framework and always allow us to form a generative SDE, necessary for the application of AYS.
} %
Looking forward, there are promising avenues for future research, including extending this framework to label- or text-conditional schedule optimization and applying it to single-step higher-order ODE solvers, such as Heun or Runge-Kutta methods. %

\section*{Broader Impact}
Diffusion models have evolved into a powerful and highly expressive generative modeling framework. Our novel method fundamentally advances diffusion models and accelerates their sampling. Faster synthesis can reduce diffusion models' inference compute demands, thereby decreasing their energy footprint, and it is also important for real-time applications. However, our approach is broadly applicable and its societal impact therefore depends on the specific domain and where and how the accelerated models are used. In our work, we validate the proposed techniques in the context of complex image and video synthesis, which have important content creation applications and can, for instance, improve the artistic workflow of digital artists and democratize creative expression. However, deep generative models like diffusion models can also be used to produce deceptive imagery and videos, as discussed, for instance, in \citet{vaccari2020deepfakes,nguyen2021deep,mirsky2021deepfakesurvey}. Therefore, they need to be used with an abundance of caution.

\bibliography{refs}
\bibliographystyle{styles/icml2024}

\newpage
\appendix
\onecolumn
\renewcommand\ptctitle{Appendix}
\addcontentsline{toc}{section}{Appendix} %
\part{} %
\parttoc %
\newpage
\section{Theoretical Details}

\subsection{Optimal Schedule for Isotropic Gaussian} \label{appendix:optimal_gaussian}
In the simple Gaussian setting where $p(\rvx) = \gN(\mathbf{0}, c^2 \rmI_{d\times d})$, the score after diffusion for time $t$ can be analytically calculated as follows
\begin{equation} \label{eq:gaussian_score}
    \nabla_\rvx \log p(\rvx, t) = -\frac{\rvx}{c^2 + t^2}.
\end{equation}

Using this score, we can derive what 1 step of forward Euler will be, when going from noise level $b$ to $a$ using the probability flow ODE (see \Cref{eq:edm_forward}) as follows
\begin{equation}
     \hat{\rvx}_a = \rvx_b + (a - b) \times \frac{b}{b^2 + c^2} \rvx_b = (\frac{ab + c^2}{b^2 + c^2}) \rvx_b.
\end{equation}
Assuming that $\rvx_b \sim \gN(\mathbf{0}, (c^2 + b^2)\rmI_{d\times d})$, the distribution obtained from forward Euler will be $\Hat{\rvx}_a \sim \gN(\mathbf{0}, (\frac{ab + c^2}{b^2 + c^2})^2 \times (b^2 + c^2)\rmI_{d\times d})$. 

This can be easily extended to $n$ steps. Assuming that $a = t_0 < t_1 < \dots < t_n = b$ is the schedule that is used to perform $n$ steps from noise level $b$ to $a$. Letting $\hat{\rvx}_a$ be the output, we have
\begin{equation}
    \Hat{\rvx}_a = \prod_{i=1}^{n} \left(\frac{t_{i-1}t_{i} + c^2}{t_{i}^2 + c^2}\right) \times \rvx_b
\end{equation}
and similarly we will have $\Hat{\rvx}_a \sim \gN(\mathbf{0}, \prod_{i=1}^{n} (\frac{t_{i-1}t_{i} + c^2}{t_{i}^2 + c^2})^2 \times (b^2 + c^2)\rmI_{d\times d})$. 

We're looking to find the optimal values of $t_i$ such that the KL-divergence between $\Hat{\rvx}_a$ and $\rvx_a$ is minimized. Since both distributions are Gaussian, the KL-divergence has a closed form:
\begin{align}
    \KL (p(x_a) \Vert p(\Hat{\rvx}_a)) &= \KL \left(\gN(\mathbf{0}, (t_0^2 + c^2) \rmI_{d\times d}), \ \  \gN(\mathbf{0}, \prod_{i=1}^{n} (\frac{t_{i-1}t_{i} + c^2}{t_{i}^2 + c^2})^2 \times (t_n^2 + c^2)\rmI_{d\times d})\right) \nonumber \\
    &= \frac{1}{2} \left[ \log \frac{|\Sigma_2|}{|\Sigma_1|} - d + tr(\Sigma_2^{-1} \Sigma_1) + (\mu_2 - \mu_1)^T \Sigma_2^{-1} (\mu_2 - \mu_1)\right] \nonumber  \\
    &= \frac{1}{2} \left[ d \times \log \frac{1}{f(t_0, \dots, t_n)} - d + d \times f(t_0, \dots, t_n) + 0 \right] \nonumber  \\
    &= \frac{d}{2} \left(-\log f(t_0, \dots, t_n) + f(t_0, \dots, t_n) - 1 \right)
\end{align}
where 
\begin{equation*}
    f(t_0, \dots, t_n) = \frac{(t_0^2 + c^2)(t_1^2 + c^2)^2(t_2^2 + c^2)^2\dots (t_{n-1}^2 + c^2)^2 (t_n^2 + c^2)}{(t_0 t_1 + c^2)^2 (t_1 t_2 + c^2)^2\dots (t_{n-1}t_{n} + c^2)^2}.
\end{equation*}

To minimize this expression w.r.t. all $t_i$ values, the partial derivatives of the expression must be zero. Formally, for $i \in [1, n-1]$ we have
\begin{equation}
    \frac{\partial}{\partial t_i} \KL (p(\rvx_a) \Vert p(\Hat{\rvx}_a)) = \frac{d}{2} \times \frac{\partial}{\partial t_i} f(t_0, \dots, t_n) \times \left(
        1 - \frac{1}{f(t_0, \dots, t_n)}
    \right).
\end{equation}
Using the Cauchy–Schwarz inequalities we have
\begin{equation}
\begin{rcases}
    (t_0^2 + c^2)(t_1^2 + c^2) > (t_0t_1 + c^2)^2 \\
    \dots \\
    (t_{n-1}^2 + c^2)(t_{n}^2 + c^2) > (t_{n-1}t_{n} + c^2)^2 \\
\end{rcases} \Rightarrow f(t_0, \dots, t_n) > 1 \Rightarrow \left(1 - \frac{1}{f(t_0, \dots, t_n)}\right) > 0,
\end{equation}
where the inequalities are strict because of the assumption that all $t_i$ are distinct. Therefore the partial derivative of $f$ w.r.t. all $t_i$ must be zero $\frac{\partial}{\partial t_i} f(t_0, \dots, t_n) = 0$.

Computing this partial derivative gives
\begin{align*}
    \frac{\partial}{\partial t_i} f(t_0, \dots, t_n) = 0 &\Rightarrow \frac{\partial}{\partial t_i} \left( \frac{(t_i^2 + c^2)^2}{(t_{i-1}t_{i} + c^2)^2(t_{i}t_{i+1} + c^2)^2}\right) = 0 \\
    &\Rightarrow \frac{\partial}{\partial t_i} \left( \frac{(t_i^2 + c^2)}{(t_{i-1}t_{i} + c^2)(t_{i}t_{i+1} + c^2)}\right) = 0 \\
\end{align*}
\begin{gather}
    \Rightarrow (2t_i)(t_{i-1}t_{i} + c^2)(t_{i}t_{i+1} + c^2) = (t_i^2 + c^2)(2t_it_{i-1}t_{i+1} + c^2(t_{i-1} + t_{i+1})) \nonumber \\
    \Rightarrow t_i^2 (t_{i-1} + t_{i+1}) + 2m(c^2 - t_{i-1}t_{i+1}) - (t_{i-1} + t_{i+1})c^2 = 0 \nonumber \\
    \Rightarrow t_i = \frac{(t_{i-1}t_{i+1} - c^2) + \sqrt{(t_{i-1}^2 + c^2)(t_{i+1}^2 + c^2)}}{t_{i-1} + t_{i+1}}. \label{eq:kl_optimal}
\end{gather}
This equation can be simplified and written as the following:
\begin{equation}
    \alpha_{i-1} \coloneqq \arctan (t_{i-1} /  c), \ \  
    \alpha_{i + 1} \coloneqq \arctan ( t_{i+1} / c ) 
    \Rightarrow t_i = 
    c \tan \left(\frac{\alpha_{i-1} + \alpha_{i+1}}{2}\right). 
\end{equation}

Using this result, the values of $\arctan(t_i / c)$ must be linear in the optimal schedule, which concludes the proof.

\subsection{Deriving the KL-Divergence Upper Bound} \label{appendix:proving_klub}
To derive \Cref{theorem:klub}, we borrowed the argument from Section 5.1 of \cite{Chen2022SamplingIA}. 

As a reminder, the following two SDEs are considered
\begin{equation*}
    \begin{cases}
        \text{SDE 1}: & d\rvx_t = \rvf_1(\rvx_{0 \to t}, t)dt + g(t) d\rvw_t \\
        \text{SDE 2}: & d\rvx_t = \rvf_2(\rvx_{0 \to t}, t)dt + g(t) d\rvw_t \\
    \end{cases}
\end{equation*}
where $\rvx_{0 \to t}$ represents the entire path from the start ($t=0$) to the current time $t$ (this formulation is useful for multi-step methods that benefit from having access to the history). 

We start with some notations. When applying Girsanov's theorem, it is convenient to think about a single stochastic process $(\rvx_t)_{t \in [0, T]}$ and to consider different measures over the path space $\gC ([0, T]; \sR^d)$. A stochastic process can be viewed as a function from sample space to path space, i.e. $\rvx(\omega): \Omega \to \gC ([0, T]; \sR^d)$. 

We define two measures over the path space:
\begin{itemize}
    \item $Q_{paths}$, under which $(\rvx_t)_{t \in [0, T]}$ has the law of SDE 1,
    \item $P_{paths}$, under which $(\rvx_t)_{t \in [0, T]}$ has the law of SDE 2.
\end{itemize}

Assume that $\rvb_t = \frac{\rvf_2(\rvx_{0 \to t}, t) - \rvf_1(\rvx_{0 \to t}, t)}{g(t)}$ and let $\rmB_t$ be a Brownian motion under $Q_{paths}$. 
Let $\gE \coloneqq \exp\left(\int_0^T \rvb_s d\rmB_s - \frac{1}{2}\int_0^T ||\rvb_s||^2 ds\right)$. 
According to \cite{Chen2022SamplingIA}, under mild regularity constraints, $\gE$ is a random variable such that $\E_{Q_{paths}} [\gE] = 1$.
Therefore, we can define $P'_{paths}$ to be a measure that satisfies $dP'_{paths}=\gE dQ_{paths}$. According to theorem 8 of \cite{Chen2022SamplingIA}, under $P'_{paths}$ the process $t \to B_t - \int_{0}^{t} b_s ds$ is a Brownian motion, which we will call $\bm{\beta}_t$. We can rewrite this as
\begin{equation} \label{eq:klub_eq1}
    d\rmB_t = \rvb_t dt + d\bm{\beta}_t.
\end{equation}

Using the definition of $Q_{paths}$, we know that
\begin{equation} \label{eq:klub_eq2}
    d\rvx_t = \rvf_1(\rvx_{0 \to t}, t) dt + g(t) d\mathbf{B}_t, \quad \rvx_0 \sim p(\rvx).
\end{equation}

Since $P'_{paths}$ is absolutely continuous with respect to $Q_{paths}$, i.e. $P'_{paths} \ll Q_{paths}$, this equation will also hold under $P'_{paths}$. Plugging \Cref{eq:klub_eq1} into the above, we can conclude that under $P'_{paths}$ we have
\begin{align}
    d\rvx_t &= \rvf_1(\rvx_{0 \to t}, t) dt + g(t) \left( \rvb_t dt + d\bm{\beta}_t \right) \\
    &= \rvf_2(\rvx_{0 \to t}, t) dt + g(t) d\bm{\beta}_t, \quad \rvx_0 \sim p(\rvx).
\end{align}
Since $\bm{\beta}_t$ is a Brownian motion under $P'_{paths}$, this becomes the exact same as SDE 2. Therefore $P'_{paths} = P_{paths}$. In other words 
\begin{equation}
    \frac{d P_{paths}}{d Q_{paths}} = \gE.
\end{equation}

Using this expression in the KL-divergence we have
\begin{align*}
    \KL (Q_{paths} || P_{paths}) &= \E_{Q_{paths}} \log \frac{dQ_{paths}}{dP_{paths}} \\
    &= -\E_{Q_{paths}} \log \gE \\
    &= \E_{Q_{paths}} \left( -\int_0^T \rvb_s d\rmB_s + \frac{1}{2} \int_0^T ||\rvb_s||^2 ds \right) \\
    &\stackrel{(i)}{=} \E_{Q_{paths}} \left( \frac{1}{2} \int_0^T ||\rvb_s||^2 ds \right) \\
    &= \frac{1}{2} \E_{Q_{paths}} \left( \int_0^T \frac{||\rvf_1(\rvx_{0 \to s},s) - \rvf_2(\rvx_{0 \to s}, s)||^2}{g(s)^2} ds \right),
\end{align*}
where $(i)$ is due to the martingale property of Ito integrals.

Since $Q, P$ are marginals of $Q_{paths}$ and $P_{paths}$ at time $t=T$, by the data processing inequality, the KL-divergence $\KL(Q||P)$ is upper-bounded by the KL-divergence $\KL(Q_{paths} || P_{paths})$ which concludes the proof.

\subsection{Early Stopping is a Necessity} \label{appendix:early_stopping_needed}
In this section, we prove that the schedule that minimizes the $\mathrm{KLUB}$ isn't necessarily going to minimize the KL-divergence too. We will do this by once again considering the simple Gaussian case where $p(\rvx) \sim \gN(\mathbf{0}, c^2 \rmI_{d \times d})$. 

The proof is by contradiction. Let's assume that the schedule minimizing the $\mathrm{KLUB}$ must also always minimize the KL as well. We'll consider the family of Extended Reverse-Time SDEs (ERSDEs introduced in \citet{ersde}) that have $h(t) = \lambda \times \sqrt{2t}$ for some constant $\lambda$. The SDE formulation for these will be
\begin{align*}
    d\rvx_t &= -(\lambda^2 + 1) t \nabla_x \log p(\rvx_t, t) \ \ dt + \lambda \sqrt{2t} \ \ d\rvw_t\\
    \Rightarrow d\rvx_t &= (\lambda^2 + 1) \left(\frac{\rvx_t - \rvx_\theta(\rvx_t, t)}{t}\right) \ \ dt + \lambda \sqrt{2t} \ \ d\rvw_t \\
    \Rightarrow d\rvx_t &= (\lambda^2 + 1) \frac{\rvx_t}{t} - (\lambda^2 + 1) \left(\frac{\rvx_\theta(\rvx_t, t)}{t}\right) \ \ dt + \lambda \sqrt{2t} \ \ d\rvw_t, 
\end{align*}
and \citet{ersde} have shown that these SDEs share the same marginals as the original reverse-time SDE and probability flow ODE. 

Assuming we use a first-order approximation for $\rvx_\theta$ as our solver, the $\mathrm{KLUB}$ of this solver will be
\begin{align}
    \mathrm{KLUB} &= \E_{\substack{\rvx_t \sim p_t \\ \rvx_{t_{next}} \sim p_{t_{next} | t}}} \int_{t_{min}}^{t_{max}} \frac{||(\lambda^2 + 1)/t \times \rvx_\theta(\rvx_t, t) - (\lambda^2 + 1)/t \times \rvx_\theta(\rvx_{t_{next}}, t_{next})||^2}{\lambda^2 \times 2t} dt \\
    &= \frac{(\lambda^2 + 1)^2}{2\lambda^2} \int_{t_{min}}^{t_{max}} \frac{1}{t^3} \E_{\substack{\rvx_t \sim p_t \\ \rvx_{t_{next}} \sim p_{t_{next} | t}}} ||\rvx_\theta(\rvx_t, t) - \rvx_\theta(\rvx_{t_{next}}, t_{next})||^2 dt \label{eq:simple_klub_early_stop},
\end{align}
where $t_{next}$ is the smallest value in the sampling schedule larger than $t$. 
Since all the terms containing $\lambda$ only appear as a constant outside the integral, the schedule that minimizes the $\mathrm{KLUB}$ is the same for all $\lambda > 0$. As such, using our initial assumption, that schedule is KL-minimizing for all these solvers as well.

Since every term in the $\mathrm{KLUB}$ can be found analytically for this Gaussian example, we can derive exactly what the $\mathrm{KLUB}$-optimal schedule will be. To do this, we start by noting that
\begin{gather*}
    -\frac{\rvx}{t^2 + c^2} = \nabla_\rvx \log p(\rvx, t) = \frac{\rvx_\theta(\rvx, t) - \rvx}{t^2}  \\
    \Rightarrow \rvx_\theta(\rvx, t) = \frac{c^2}{t^2 + c^2} \rvx
\end{gather*}
We can use this to calculate the expectations inside the integral of \Cref{eq:simple_klub_early_stop} as follows %
\begin{align*}
    \E_{\substack{\rvx_t \sim p_t \\ \rvx_{t_{next}} \sim p_{t_{next} | t}}} ||\rvx_\theta(\rvx_t, t) - \rvx_\theta(\rvx_{t_{next}}, t_{next})||^2 &= \E_{\substack{\rvx_t \sim \gN(0, (t^2 + c^2)\rmI) \\ \rvx_{t_{next}} \sim \gN (\rvx_{t}, \sqrt{t_{next}^2 - t^2} \rmI)}} ||\rvx_\theta(\rvx_t, t) - \rvx_\theta(\rvx_{t_{next}}, t_{next})||^2 \\
    &= \E_{\substack{\rvx_t \sim \gN(0, (t^2 + c^2)\rmI) \\ \bm{\epsilon} \sim \gN(0, \rmI)}} 
    ||\rvx_\theta(\rvx_t, t) - \rvx_\theta(\rvx_{t} + \sqrt{t_{next}^2 - t^2} \bm{\epsilon}, t_{next})||^2 \\
    &= \E_{\substack{\rvx_t \sim \gN(0, (t^2 + c^2)\rmI) \\ \bm{\epsilon} \sim \gN(0, \rmI)}} 
    \left\lVert\frac{c^2}{t^2 + c^2} \rvx_t - \frac{c^2}{t_{next}^2 + c^2} (\rvx_t + \sqrt{t_{next}^2 - t^2} \bm{\epsilon})\right\rVert^2 \\
    &= c^4 \E_{\substack{\rvx_t \sim N(0, (t^2 + c^2)\rmI) \\ \bm{\epsilon} \sim \gN(0, \rmI)}} 
    \left\lVert \left(\frac{1}{t^2 + c^2} - \frac{1}{t_{next}^2 + c^2}\right)\rvx_t - \frac{\sqrt{t_{next}^2 - t^2}}{t_{next}^2 + c^2} \bm{\epsilon} \right\rVert^2 \\
    &= c^4 \times d \times \left[ \left(\frac{1}{t^2 + c^2} - \frac{1}{t_{next}^2 + c^2}\right)^2 (t^2 + c^2) + \frac{t_{next}^2 - t^2}{(t_{next}^2 + c^2)^2} \right] \\
    &= c^4 \times d \times \left( \frac{1}{t^2 + c^2} - \frac{1}{t_{next}^2 + c^2} \right), 
\end{align*}
where we used the fact that $\rvx_t, \bm{\epsilon}$ are independent random variables, and $\bm{\epsilon}$ has zero mean. 

Assuming the sampling schedule is $a = t_0 < t_1 < \dots < t_n = b$ and plugging this into \Cref{eq:simple_klub_early_stop} we have
\begin{align*}
    \mathrm{KLUB} &\propto  \int_{t_{min}}^{t_{max}} \frac{1}{t^3} \left( \frac{1}{t^2 + c^2} - \frac{1}{t_{next}^2 + c^2} \right) dt \\
    &\propto \sum_{i=1}^{n} \left[ -\frac{c^2 t_{i}^2}{c^2 + t_i^2} \left( \frac{1}{t_i^2} - \frac{1}{t_{i-1}^2} \right) + \log \left(\frac{t_i^2 + c^2}{t_{i-1}^2 + c^2}\right) - \log \left(\frac{t_i^2}{t_{i-1}^2}\right) \right] \\
    &= \log \left(\frac{b^2 + c^2}{a^2 + c^2}\right) - \log \frac{b^2}{a^2} + c^2 \sum_{i=1}^{n} \frac{t_i^2 - t_{i-1}^2}{(c^2 + t_i^2) \times t_{i-1}^2}.
\end{align*}

To derive the optimal $t_i$ values, the partial derivative of the expression above w.r.t. each $t_i$ must be zero. Writing the partial derivative w.r.t. $t_i$ and setting it to zero and simplifying yields
\begin{equation}
    \frac{c^2}{t_i^2} = \frac{\sqrt{(t_{i-1}^2 + c^2)(t_{i+1}^2 + c^2)} - t_{i-1}t_{i+1}}{t_{i-1}t_{i+1}}.
\end{equation}

Now, we want to figure out what these solvers will look like when $\lambda \to 0$. Deriving what a single step of the solver from noise level $b$ to $a$ will be gives the following update rule
\begin{equation}
    \rvx_a = \left(\frac{a}{b}\right)^{\lambda^2 + 1} \rvx_b + \rvx_\theta(\rvx_b, b) \left( 1 - (\frac{a}{b})^{\lambda^2 + 1}\right) + a\sqrt{1 - (\frac{a}{b})^{2\lambda^2}} \ \rvz_b,
    \label{eq:klub_optimal}
\end{equation}
where $\rvz_b \sim \gN(0, \rmI)$. In the limit when $\lambda \to 0$, this reduces to one step of forward Euler on the probability flow ODE. 
Using our assumption that the $\mathrm{KLUB}$-optimal schedule is KL-optimal, when $\lambda \to 0$ the same sampling schedule is KL-optimal for all $\lambda$, and the update rule gets closer and closer to the forward Euler update rule. As such, this schedule must also be KL-optimal for forward Euler as well. 

At this point, we have explicit expression for both the $\mathrm{KLUB}$ optimal schedule and forward Eulers's KL-optimal schedules from \Cref{eq:klub_optimal,eq:kl_optimal}. 
\begin{align*}
\text{KL optimal schedule}: & \ \ t_i = \frac{(t_{i-1}t_{i+1} - c^2) + \sqrt{(t_{i-1}^2 + c^2)(t_{i+1}^2 + c^2)}}{t_{i-1} + t_{i+1}}\\
\text{KLUB optimal schedule}: & \ \  t_i = c \times \sqrt{\frac{t_{i-1}t_{i+1}}{\sqrt{(t_{i-1}^2 + c^2)(t_{i+1}^2 + c^2)} - t_{i-1}t_{i+1}}}\\
\end{align*}
A simple comparison between these two equations makes it clear they are not the same, which is a contradiction, disproving our initial assumption.

Therefore, using the $\mathrm{KLUB}$ objective to optimize the schedule does not translate directly into minimizing the mismatch between final output distributions. 
This is expected, given that the objective measures the divergence in the path distributions of the two SDEs, which is an upper bound for the mismatch between output distributions. 
As a result, optimizing the schedule with this objective focuses on aligning the trajectories of the two SDEs, rather than their end states. 

Empirically, we've found that the heuristic schedules commonly in use are extremely sub-optimal, affecting both output and path distributions. 
Optimizing these schedules using the $\mathrm{KLUB}$ objective aligns the solver's paths with the true paths of the exact SDE. Initially, this alignment process also brings the output distributions closer together. However, after a certain point, the process begins to favor the alignment of intermediate noised distributions at the expense of the final output distributions, leading to more closely aligned paths. 
Therefore, early stopping must be used to prevent this from happening.

\subsection{Exact Error Analysis} \label{appendix:error_analysis}
In \Cref{sec:KLUB} we assumed the learnt model is very close to the ideal denoiser, which let us approximate $P^{\text{true paths}}$ with the path distribution from the forward noising process. 
In this section, we provide a detailed analysis without that assumption, deriving an accurate $\mathrm{KLUB}$ that includes the model's approximation error. We start with the following assumption: 
\begin{assumption}[score estimation error]
    Letting $D^*$ be the ideal denoiser, for all $t \in [t_{min}, t_{max}]$ the score estimation error is bounded: 
    \begin{equation*}
        \E_{\rvx_t \sim p'(\rvx, t)} \left\lVert  {\color{Green} \rmD^*}\left({\frac{\rvx_{t}}{s(t)}, \sigma(t)}\right) -  {\color{Red} \rmD_\theta}\left({\frac{\rvx_t}{s(t)}, \sigma(t)}\right) \right\rVert^2 \leq \epsilon^2.
    \end{equation*}
\end{assumption}

Using this, we apply \Cref{theorem:klub} to calculate the $\mathrm{KLUB}$ between the true reverse SDE, which contains the ideal denoiser $D^*$, and the discretized linear SDE. Let $P_{t_i \to t_{i-1}}^{\text{exact}}$ represent the path distributions of the exact revere-time SDE, which matches the paths of the forward noising process. Then we have:
\begin{equation*}
\begin{split}
    \KL (P_{t_{i} \to t_{i-1}}^{\textrm{exact}} \Vert P_{t_{i} \to t_{i-1}}^{\textrm{disc}}) &\leq 
    2 \times \E_{P_{t_{i} \to t_{i-1}}^{\text{exact}}} \int_{t_{i-1}}^{t_{i}} \frac{s(t)^2 \dot{\sigma}(t)}{\sigma(t)^3} \left\lVert {\color{Green} \rmD^*}\left({\color{Green} \frac{\rvx_t}{s(t)}, \sigma(t)}\right) -  {\color{Red} \rmD_\theta}\left({\color{Red} \frac{\rvx_{t_{i}}}{s(t_{i})}, \sigma(t_{i})}\right) \right\rVert^2 dt \\
    &= 2 \times  \int_{t_{i-1}}^{t_{i}} \frac{s(t)^2 \dot{\sigma}(t)}{\sigma(t)^3} \times  \E_{\substack{\rvx_t \sim p'(\rvx, t) \\ \rvx_{t_{i}} \sim p'(\rvx_{t_{i}} | \rvx_t)}} \left( \left\lVert {\color{Green} \rmD^*}\left({\color{Green} \frac{\rvx_t}{s(t)}, \sigma(t)}\right) -  {\color{Red} \rmD_\theta}\left({\color{Red} \frac{\rvx_{t_{i}}}{s(t_{i})}, \sigma(t_{i})}\right) \right\rVert^2 \right) dt \\
\end{split}
\end{equation*}
\begin{equation*}
\begin{gathered}
    \leq 4 \times  \int_{t_{i-1}}^{t_{i}} \frac{s(t)^2 \dot{\sigma}(t)}{\sigma(t)^3} \times  \E_{\substack{\rvx_t \sim p'(\rvx, t) \\ \rvx_{t_{i}} \sim p'(\rvx_{t_{i}} | \rvx_t)}} \left( \left\lVert {\color{Green} \rmD^*}\left({\color{Green} \frac{\rvx_t}{s(t)}, \sigma(t)}\right) - {\color{Red} \rmD_\theta}\left({\color{Green} \frac{\rvx_t}{s(t)}, \sigma(t)}\right) \right\rVert^2 + \left\lVert {\color{Red} \rmD_\theta}\left({\color{Green} \frac{\rvx_t}{s(t)}, \sigma(t)}\right) - {\color{Red} \rmD_\theta}\left({\color{Red} \frac{\rvx_{t_{i}}}{s(t_{i})}, \sigma(t_{i})}\right) \right\rVert^2 \right) dt \\
    \leq 4 \times  \int_{t_{i-1}}^{t_{i}} \frac{s(t)^2 \dot{\sigma}(t)}{\sigma(t)^3} \times  \E_{\substack{\rvx_t \sim p'(\rvx, t) \\ \rvx_{t_{i}} \sim p'(\rvx_{t_{i}} | \rvx_t)}} \left( \epsilon^2 + \left\lVert {\color{Red} \rmD_\theta}\left({\color{Green} \frac{\rvx_t}{s(t)}, \sigma(t)}\right) - {\color{Red} \rmD_\theta}\left({\color{Red} \frac{\rvx_{t_{i}}}{s(t_{i})}, \sigma(t_{i})}\right) \right\rVert^2 \right) dt \\
    = \underbrace{\vphantom{
        4 \times  \int_{t_{i-1}}^{t_{i}} \frac{s(t)^2 \dot{\sigma}(t)}{\sigma(t)^3} \times  \E_{\substack{\rvx_t \sim p'(\rvx, t) \\ \rvx_{t_{i}} \sim p'(\rvx_{t_{i}} | \rvx_t)}} \left( \left\lVert {\color{Red} \rmD_\theta}\left({\color{Green} \frac{\rvx_t}{s(t)}, \sigma(t)}\right) - {\color{Red} \rmD_\theta}\left({\color{Red} \frac{\rvx_{t_{i}}}{s(t_{i})}, \sigma(t_{i})}\right) \right\rVert^2 \right) dt
    } O(\epsilon^2)}_{\text{Approximation error}} +  \underbrace{4 \times  \int_{t_{i-1}}^{t_{i}} \frac{s(t)^2 \dot{\sigma}(t)}{\sigma(t)^3} \times  \E_{\substack{\rvx_t \sim p'(\rvx, t) \\ \rvx_{t_{i}} \sim p'(\rvx_{t_{i}} | \rvx_t)}} \left( \left\lVert {\color{Red} \rmD_\theta}\left({\color{Green} \frac{\rvx_t}{s(t)}, \sigma(t)}\right) - {\color{Red} \rmD_\theta}\left({\color{Red} \frac{\rvx_{t_{i}}}{s(t_{i})}, \sigma(t_{i})}\right) \right\rVert^2 \right) dt}_{\text{Discretization error}}.
\end{gathered}
\end{equation*}
This means in the exact case, the $\mathrm{KLUB}$ can be broken into 2 parts, namely an approximation error and a discretization error. The approximation error relies solely on the model, and can only be improved by training the model further.  
Therefore, we can ignore it and focus on minimizing the discretization error, which leads to the $\mathrm{KLUB}$ objective derived in \Cref{sec:KLUB}.

\section{Experiment Details}

\subsection{Practical Implementation Details} \label{appendix:importance_sampling}
As mentioned in \Cref{sec:importance_sampling}, in practice to optimize a schedule for a given model and dataset, first a 10-step schedule $(t_0, t_1, \dots, t_{10})$ is initialized using one of the baseline hand-crafted schedules. For continuous-time models, we initialize the schedule according to the EDM scheme, and for discrete-time models, time-uniform initialization is used. 

\rnew{
Afterwards, the schedule is optimized in a hierarchical manner. 
This is done by first optimizing all the 9 intermediate points $(t_1, t_2, \dots, t_9)$ of the schedule iteratively and using an early-stopping mechanism to avoid over-fitting. Perceptual image quality metrics or even manual inspection can be used as proxies to determine the stopping point of the optimization. 
Next, for 2 rounds, a subdivision operation is done that doubles the number of steps of the schedule, and further fine-tuning is performed on the newly added intermediate points, keeping the initial previous-round points fixed. These fine-tuning stages do not need early stopping due to the fixed ``shape'' of the schedule from the first-round optimization. 
The main reason behind using hierarchical optimization is to speed up training which is due to two factors. First, when optimizing a specific point $t_i$ of a schedule, the optimized value will always remain inside $[t_{i-1}, t_{i+1}]$. As a result, if instead of hierarchical optimization, we optimized a 40-step schedule directly, the changes of each point would be smaller (due to the tighter $[t_{i-1}, t_{i+1}]$ intervals), resulting in more iterations to converge. Secondly, after each subdivision, only half of the points of a schedule are being optimized and these points are non-adjacent, making each point’s optimization independent of the others. This allows the entire process to be parallelized. Furthermore, since during the later stages each point being optimized lies in a fixed interval (its endpoints are frozen), a very small number of iterations is required for it to converge.
}

To optimize the $i$-th element $t_i$, a number of candidate values are chosen in a neighbourhood around $t_i$, and for each the KLUB value is estimated with Monte-Carlo integration. Finally, $t_i$ is set to be the candidate with the minimum KLUB value. In practice, we select 11 candidates with the current $t_i$ value being one of them to ensure the KLUB is always decreasing. A pseudocode for this is given in \Cref{alg:klub_optim}. 
\rnew{
We also experimented with having $t_i$ being learnable parameters that are differentially optimized with respect to the KLUB loss term. We tried two different scenarios where 
$t_i$’s are all optimized simultaneously or iteratively. In our experiments, we found this approach to be extremely unstable, requiring heavy fine-tuning of hyperparameters as well as a large effective batch size to smooth the gradient estimates. The large batch size also resulted in very slow optimization. As a result, we opted for the zeroth-order optimization approach, which does not rely on noisy gradient estimates. This optimization is relatively low-dimensional, consisting of only a small set of time steps that need to be adjusted, and zeroth-order optimization can work well in such settings.
}

\begin{figure}[h]
    \centering
    \includegraphics[width=0.4\linewidth]{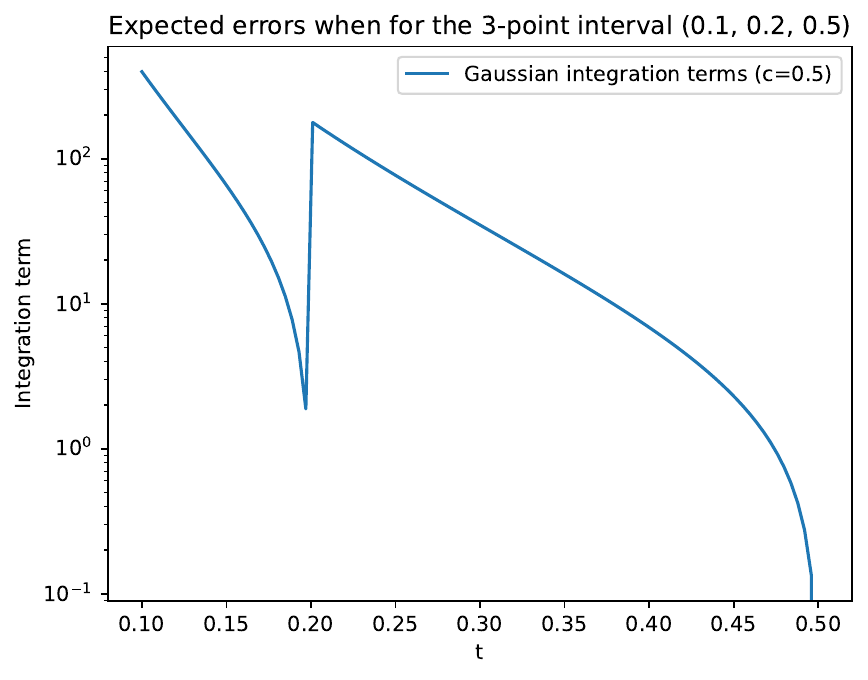}
    \caption{
        The figure illustrates the integration values of the $\mathrm{KLUB}$ in a Gaussian data setting within the interval $(0.1, 0.5)$ assuming a schedule of $(t_{i-1}, t_{i}, t_{i+1}) = (0.1, 0.2, 0.5)$, highlighting a large range of values and a discontinuity at $0.2$. %
    }
    \label{fig:gaussian_integration_vals}
\end{figure}

To perform the Monte-Carlo integration, as discussed in \Cref{sec:importance_sampling}, importance sampling is used. 
As mentioned previously, this is due to the integration values of the KLUB in \Cref{eq:klub_final} varying greatly in size. For example, we visualize these integration values for $t \in [t_{i-1}, t_{i+1}]$ where $(t_{i-1}, t_{i}, t_{i+1}) = (0.1, 0.2, 0.5)$ in \Cref{fig:gaussian_integration_vals}. The discontinuity at $t=0.2$ highlights the point where the integrand values change from $(\frac{1}{t^2 + c^2} - \frac{1}{0.2^2 + c^2})$ for $t \in (0.1, 0.2]$ to $(\frac{1}{t^2 + c^2} - \frac{1}{0.5^2 + c^2})$ for $t \in (0.2, 0.5]$. As can be seen, in this example, the values range from $[0 - 1000]$ spanning roughly 3 orders of magnitude. A pseudocode is given in \Cref{alg:klub_mc}.

In practice, we found using a subset of 8192 data samples with the time-based importance sampling to work well, and this configuration is used in all our experiments. \rnew{Although the subset of samples used for estimating the KLUB isn't comprehensive, this is large enough to capture the essence of the data set and the optimized schedules generalize to the entire dataset. One interpretation of this could be that hand-designed schedules are so far from optimal that even optimizing them with respect to a small set of samples from the distribution gives substantial improvements. Moreover, note that we are optimizing a sampling schedule consisting of only a handful of timesteps. This represents a rather low-dimensional optimization problem, which may not require a large training datasets (in contrast, for instance, to the very high-dimensional optimization problem of training a large neural network from data).}

\rnew{Finally, the optimization time needed for different models depends heavily on how resource-heavy the model is because of the Monte-Carlo integration. However, in practice, each optimization only required at most 300 iterations. In our experiments, we used RTX6000 GPUs to carry out the optimization. The FFHQ and CIFAR10 experiments required 4 GPUs for 1.5 hours. The ImageNet 256x256 and text-to-image experiments were done with 8 GPUs and took roughly 3-4 hours. Lastly, the Stable Video Diffusion experiments were done with 16 GPUs and took 6 hours. 
It is worth noting that since the majority of training time was spent on Monte-Carlo integration and forward passing through the score network, increasing the number of GPUs linearly would almost linearly decrease the amount of time spent.}

\begin{algorithm}[h]
   \caption{$\mathrm{KLUB}$ optimization with $\sigma(t) = t$ and $s(t) = 1$.}
   \label{alg:klub_optim}
\begin{algorithmic}[1]
   \STATE {\bfseries Input:} denoiser $\rmD_\theta(\rvx, \sigma)$, schedule $t_{i \in \{0, 1, \dots, n\}}$
   \REPEAT  
   \STATE Initialize $noChange = \textrm{True}$
   \FOR{$i=1$ {\bfseries to} $n-1$}
   \STATE $candidates[0, \dots, r-1] \gets$ Neighbourhood around $t_{i}$
   
   \FOR{$j=0$ {\bfseries to} $r-1$}
   \STATE $KLUB[j] \gets \textrm{EstimateKLUB}(\rmD_\theta, $ $\{t_{i-1}, candidates_j, t_{i+1}\})$
   \ENDFOR

   \STATE $\textrm{minIdx} \gets \arg\min KLUB[0, \dots, r-1]$
   \IF{$candidate_{\textrm{minIdx}} \neq t_{i}$}
   \STATE $t_{i} \gets candidate_{\textrm{minIdx}}$
   \STATE $noChange \gets \textrm{False}$
   \ENDIF
   \ENDFOR
   \UNTIL{$noChange$}
\end{algorithmic}
\end{algorithm}

\begin{algorithm}[h]
   \caption{Monte Carlo estimation of $\mathrm{KLUB}$ with $\sigma(t) = t$ and $s(t) = 1$.}
   \label{alg:klub_mc}
\begin{algorithmic}[1]
    \STATE {\bfseries Input:} denoiser $\rmD_\theta(\rvx, \sigma)$, interval points $t_{min}$, $t_{mid}$, $t_{max}$, monte carlo samples $n$
    
    \FOR{$i=1$ {\bfseries to} $n$}
    \STATE {\bfseries sample} $\rvx_0 \sim p_{data}(\rvx)$ 
    \STATE $t \leftarrow ImportanceSample(\pi, t_{min}, t_{mid}, t_{max})$ 
    \STATE $t_{upper} \gets (t < t_{mid}) \ ? \ t_{mid} : t_{max}$

    \STATE $\rvx_t \gets \rvx_0 + t \times \gN(\mathbf{0}, \rmI)$
    \STATE $\rvx_{t_{upper}} \leftarrow \rvx_t + \sqrt{t_{upper}^2 - t^2} \times \gN(\mathbf{0}, \rmI)$

    \STATE $KLUB[i] \gets ||\rmD_\theta(\rvx_t, t) - \rmD_\theta(\rvx_{t_{upper}}, t_{upper})||^2 / (\frac{1}{t^2 + c^2} - \frac{1}{t_{upper}^2 + c^2})$
    
    \ENDFOR

    \STATE $t_{upper} \leftarrow (t < t_{mid}) \ ? \ t_{mid} :$
    \STATE \hspace{\algorithmicindent}$t_{max}$

    \RETURN $\textrm{mean}(KLUB[0, \dots, n-1])$
\end{algorithmic}
\end{algorithm}

\subsection{Popular Sampling Schedules} \label{appendix:popular_schedules}
Currently, most diffusion models use one of a handful of different hand-designed sampling schedules at inference. Below we go over some of the most popular ones.

\textbf{EDM Schedule:} This schedule first introduced by \cite{edm} chooses the sampling schedule as follows:
\begin{equation*}
    \sigma(t_{i}) = (\sigma_{min}^{\frac{1}{\rho}} + (\sigma_{max}^{\frac{1}{\rho}} - \sigma_{min}^{\frac{1}{\rho}}) \times \frac{i}{n})^\rho,
\end{equation*}
where $\rho=7$ is usually used.  

\textbf{LogSNR schedule:} This schedule is a special case of EDM's schedule where $\rho=1$. Specifically:
\begin{equation*}
    \sigma(t_{i}) = (\sigma_{min} + (\sigma_{max} - \sigma_{min}) \times \frac{i}{n}).
\end{equation*}

\textbf{Time-uniform schedule:} This schedule is mainly used in discrete models. In these cases, the sampling schedule is simply:
\begin{equation*}
    t_i = \epsilon + \frac{i}{n}(1 - \epsilon).
\end{equation*}
In this case, the schedule will mimic the noise schedule with which the model was trained.

\subsection{Optimized Schedules for Large Scale Models}
We provide our optimized schedules for Stable Diffusion 1.5, SDXL, DeepFloyd-IF, and Stable Video Diffusion in \Cref{table:optimized_schedules}. The values in the table are the noise levels for the different steps i.e. $\sigma(t_i)$.

\begin{table}[H]
\centering
\caption{Optimized schedules. The values represent the noise levels of the schedule $\sigma(t_{n}), \sigma(t_{n-1}), \dots, \sigma(t_{0})$.} 
\label{table:optimized_schedules}
\resizebox{\linewidth}{!}{
\begin{tabular}{@{} lc @{}}
\toprule
                                                            & Optimized Schedules \\
\midrule
Stable Diffusion 1.5 \citep{Rombach2021HighResolutionIS}    & [14.615, 6.475, 3.861, 2.697, 1.886, 1.396, 0.963, 0.652, 0.399, 0.152, 0.029]   \\
SDXL \citep{Podell2023SDXLIL}                               & [14.615, 6.315, 3.771, 2.181, 1.342, 0.862, 0.555, 0.380, 0.234, 0.113, 0.029]   \\
DeepFloyd-IF / Stage 1 \citep{Deepfloyd}                    & [160.41, 8.081, 3.315, 1.885, 1.207, 0.785, 0.553, 0.293, 0.186, 0.030, 0.006]   \\
Stable Video Diffusion \citep{Blattmann2023StableVD}        & [700.00, 54.5, 15.886, 7.977, 4.248, 1.789, 0.981, 0.403, 0.173, 0.034, 0.002]   \\
\bottomrule
\end{tabular}
}
\end{table}

\section{Additional Results}

\subsection{Gaussian Data Extras} \label{appendix:gaussian_data_extras}
\begin{figure}[H]
    \centering
    \subfigure[$c=0.1$]{
        \includegraphics[width=0.3\textwidth]{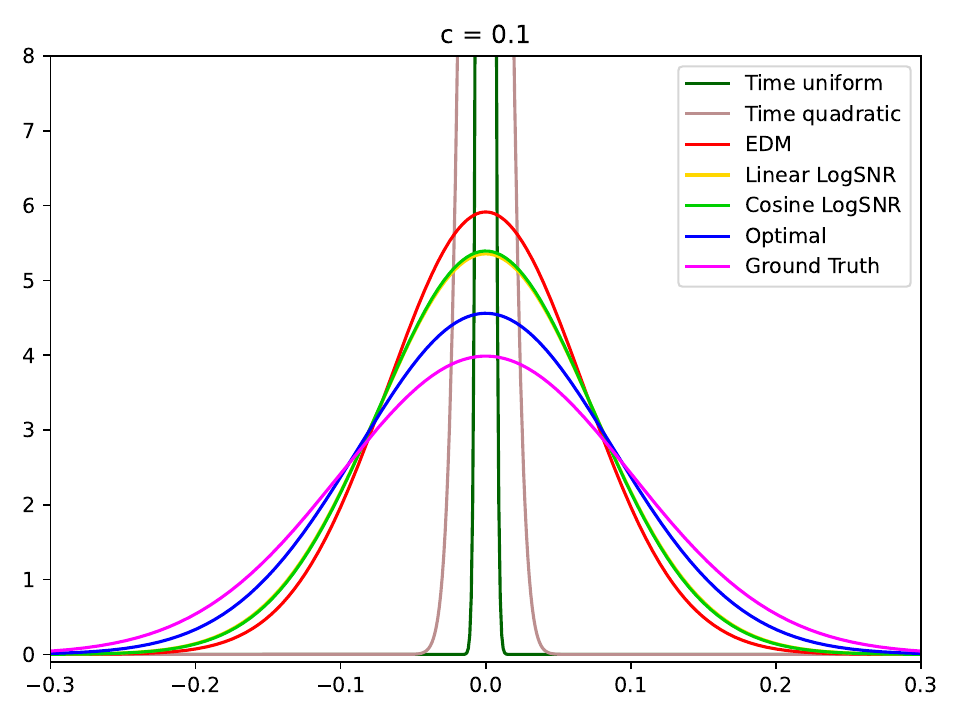}
    }%
    \subfigure[$c=0.5$]{
        \includegraphics[width=0.3\textwidth]{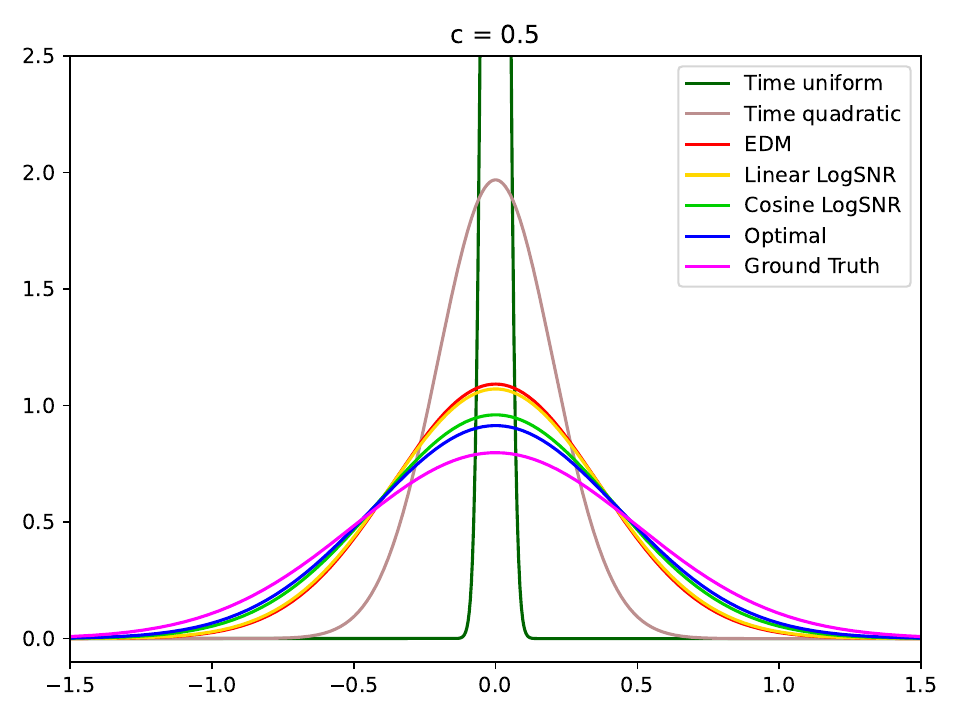}
    }%
    \subfigure[$c=1.0$]{
        \includegraphics[width=0.3\textwidth]{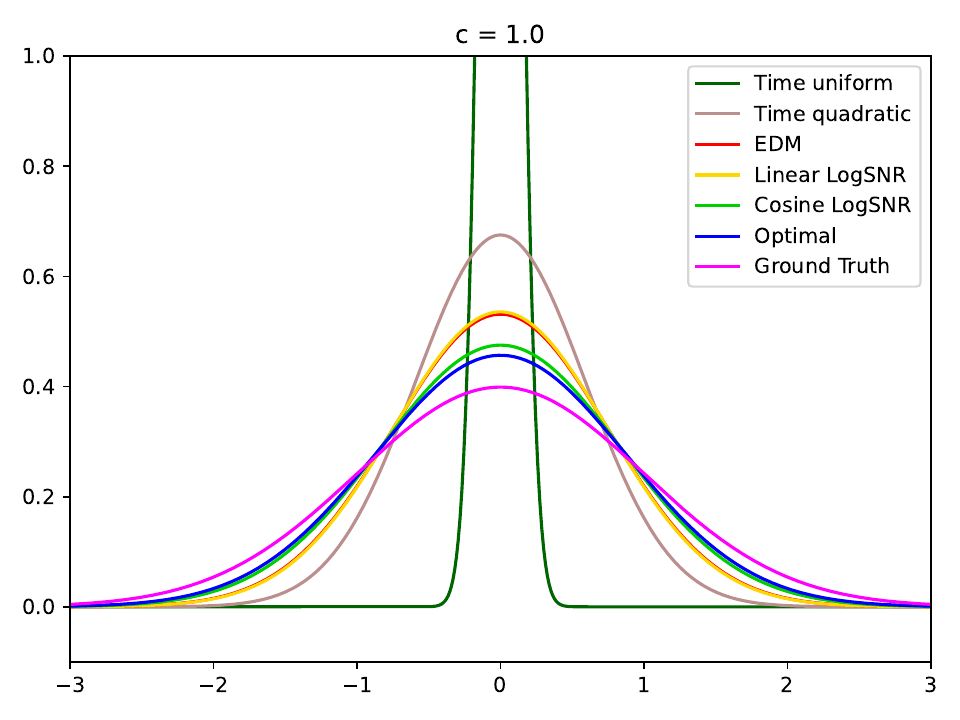}
    }
    \caption{ 
        Comparing the output distributions of using various schedules for different initial $c$ values in the Gaussian setting.
    }    
    \label{fig:simple_gaussian_output_distributions}
\end{figure}

\subsection{Extra 2D Experiments} \label{appendix:extra_2d_experiments}
\begin{figure}[H]
    \centering
    \begin{minipage}{0.25\linewidth}
        \centering
        \includegraphics[width=\linewidth]{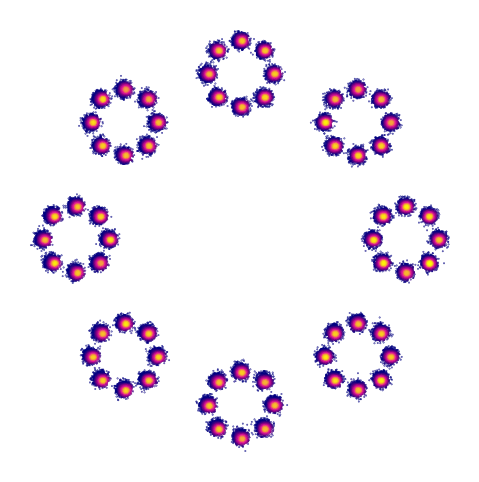} 
        \captionof{subfigure}{Ground truth}
    \end{minipage}%
    \begin{minipage}{0.25\linewidth}
        \centering
        \includegraphics[width=\linewidth]{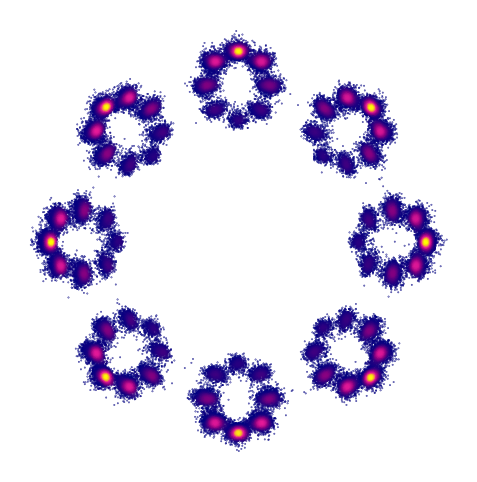} 
        \captionof{subfigure}{EDM}
    \end{minipage}%
    \begin{minipage}{0.25\linewidth}
        \centering
        \includegraphics[width=\linewidth]{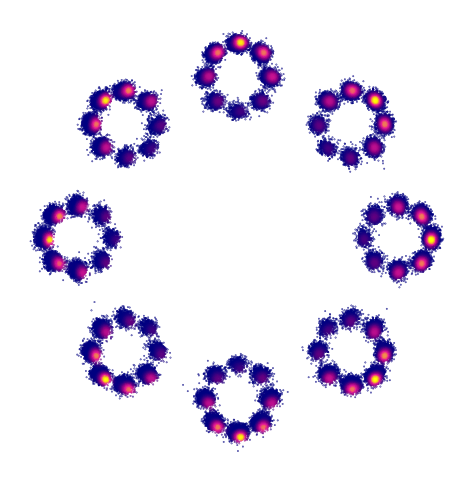} 
        \captionof{subfigure}{LogSNR}
    \end{minipage}%
    \begin{minipage}{0.25\linewidth}
        \centering
        \includegraphics[width=\linewidth]{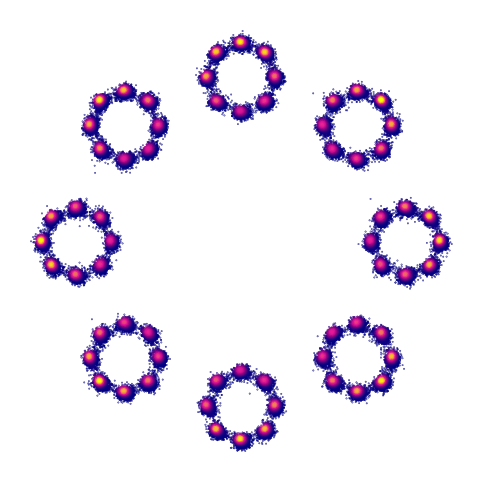} 
        \captionof{subfigure}{AYS}
    \end{minipage}
    \caption{
        \textbf{Modeling a 2D toy distribution}: (a) Ground truth samples; (b), (c), and (d) are samples generated using 10 steps of SDE-DPM-Solver++(2M) with EDM, LogSNR, and AYS schedules, respectively. Each image consists of 100,000 sampled points. 
    }    
    \label{fig:toy_8x8}
\end{figure}

\begin{figure}[h]
    \centering
    \begin{minipage}{0.25\linewidth}
        \centering
        \includegraphics[width=\linewidth]{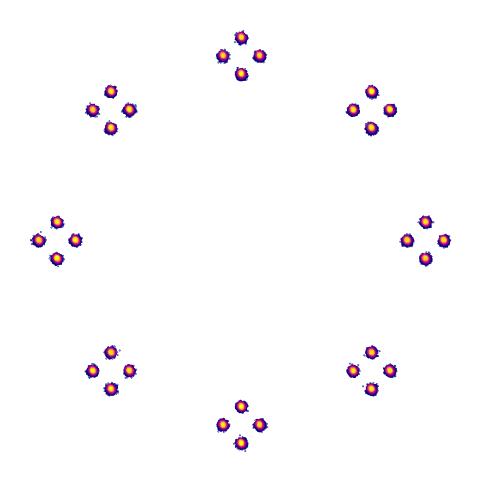} 
        \captionof{subfigure}{Ground truth}
    \end{minipage}%
    \begin{minipage}{0.25\linewidth}
        \centering
        \includegraphics[width=\linewidth]{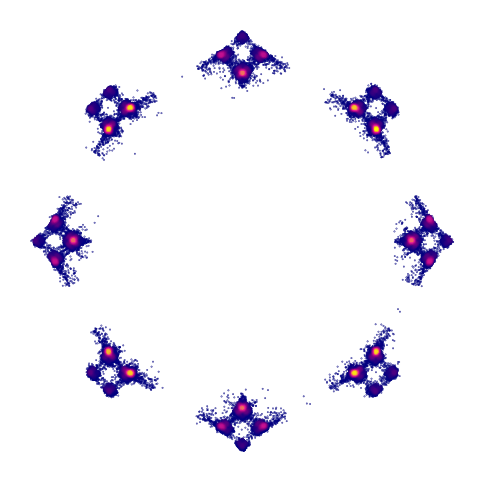} 
        \captionof{subfigure}{EDM}
    \end{minipage}%
    \begin{minipage}{0.25\linewidth}
        \centering
        \includegraphics[width=\linewidth]{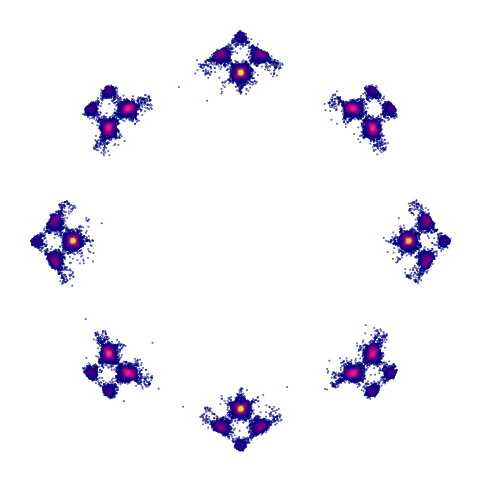} 
        \captionof{subfigure}{LogSNR}
    \end{minipage}%
    \begin{minipage}{0.25\linewidth}
        \centering
        \includegraphics[width=\linewidth]{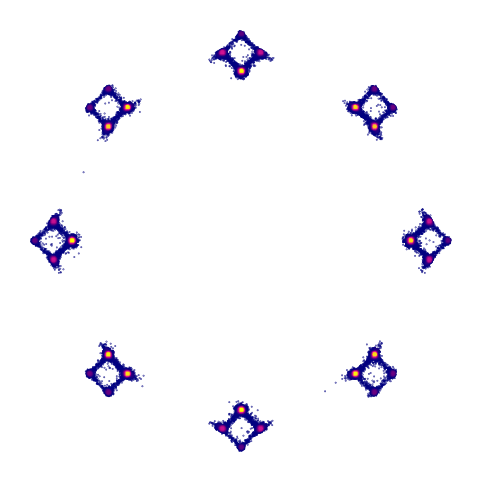} 
        \captionof{subfigure}{AYS}
    \end{minipage}
    \caption{
        \textbf{Modeling a 2D toy distribution}: (a) Ground truth samples; (b), (c), and (d) are samples generated using 7 steps of DDIM with EDM, LogSNR, and AYS schedules, respectively. Each image consists of 100,000 sampled points. 
    }    
    \label{fig:toy_8x4}
\end{figure}

\begin{figure}[h]
    \centering
    \begin{minipage}{0.25\linewidth}
        \centering
        \includegraphics[width=\linewidth]{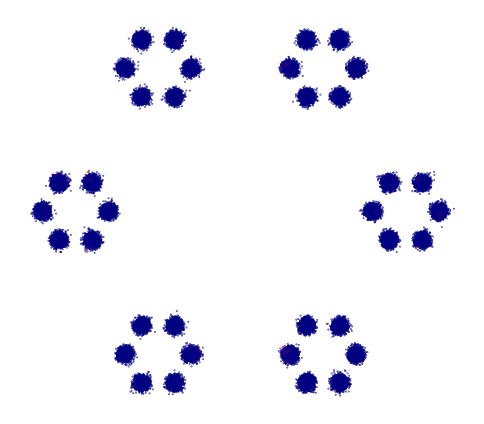} 
        \captionof{subfigure}{Ground truth}
    \end{minipage}%
    \begin{minipage}{0.25\linewidth}
        \centering
        \includegraphics[width=\linewidth]{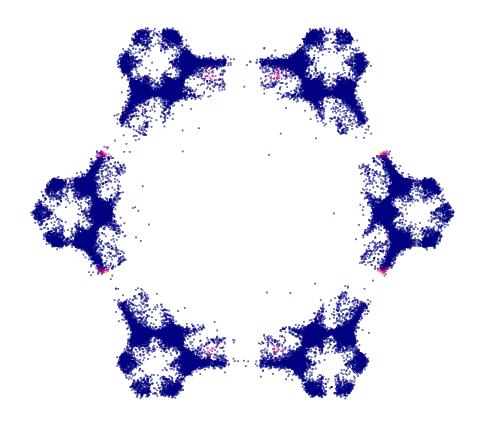} 
        \captionof{subfigure}{EDM}
    \end{minipage}%
    \begin{minipage}{0.25\linewidth}
        \centering
        \includegraphics[width=\linewidth]{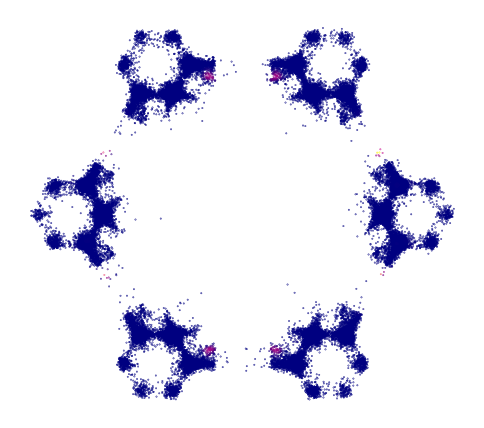} 
        \captionof{subfigure}{LogSNR}
    \end{minipage}%
    \begin{minipage}{0.25\linewidth}
        \centering
        \includegraphics[width=\linewidth]{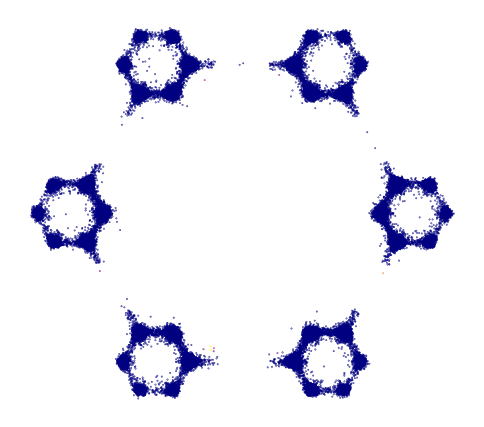} 
        \captionof{subfigure}{AYS}
    \end{minipage}
    \caption{
        \textbf{Modeling a 2D toy distribution}: (a) Ground truth samples; (b), (c), and (d) are samples generated using 6 steps of Stochastic-DDIM with EDM, LogSNR, and AYS schedules, respectively. Each image consists of 100,000 sampled points. 
    }    
    \label{fig:toy_6x6}
\end{figure}

In this section, we provide extra experiments for various 2D toy data. 
First we consider a set of datasets for which we know the ground truth score analytically, i.e. a mixture of gaussians. We consider 3 different variants of mixture of gaussians and show samples generated with various samplers using the EDM, LogSNR, and optimized schedules in \Cref{fig:toy_8x8,fig:toy_8x4,fig:toy_6x6}. 
We also report negative log-likelihoods (NLL) for these samples in \Cref{table:toy_nll}. 

\begin{figure}[H]
    \centering
    \begin{minipage}{0.25\linewidth}
        \centering
        \includegraphics[width=\linewidth]{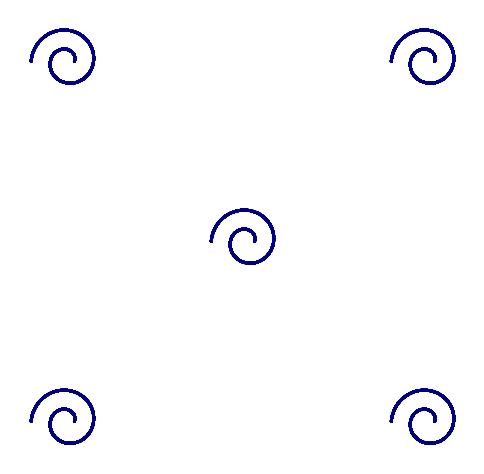} 
        \captionof{subfigure}{Ground truth}
    \end{minipage}%
    \begin{minipage}{0.25\linewidth}
        \centering
        \includegraphics[width=\linewidth]{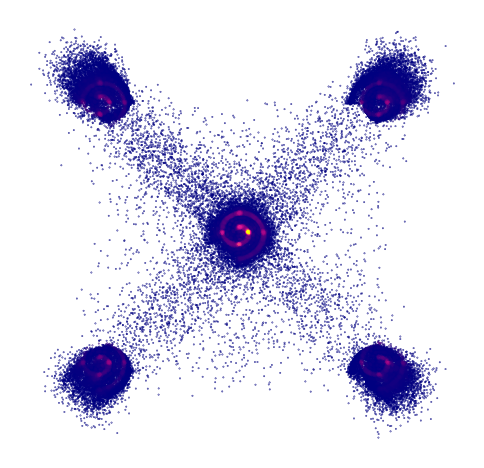} 
        \captionof{subfigure}{EDM}
    \end{minipage}%
    \begin{minipage}{0.25\linewidth}
        \centering
        \includegraphics[width=\linewidth]{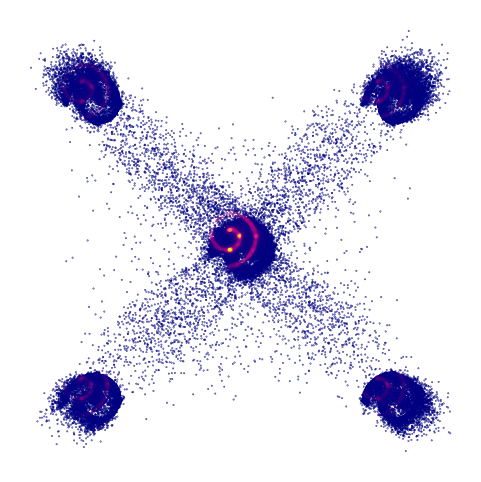} 
        \captionof{subfigure}{LogSNR}
    \end{minipage}%
    \begin{minipage}{0.25\linewidth}
        \centering
        \includegraphics[width=\linewidth]{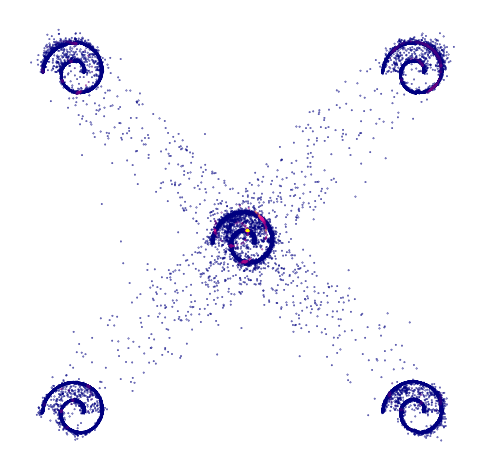} 
        \captionof{subfigure}{AYS}
    \end{minipage}%
    \caption{
        \textbf{Modeling a 2D toy distribution}: (a) Ground truth samples; (b), (c), and (d) are samples generated using 6 steps of SDE-DPM-Solver++(2M) with EDM, LogSNR, and AYS schedules, respectively. Each image consists of 100,000 sampled points. 
    }    
    \label{fig:toy_spiral}
\end{figure}

\begin{figure}[h]
    \centering
    \begin{minipage}{0.25\linewidth}
        \centering
        \includegraphics[width=\linewidth]{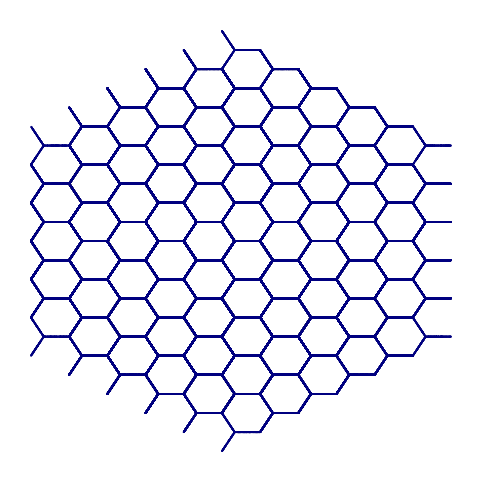} 
        \captionof{subfigure}{Ground truth}
    \end{minipage}%
    \begin{minipage}{0.25\linewidth}
        \centering
        \includegraphics[width=\linewidth]{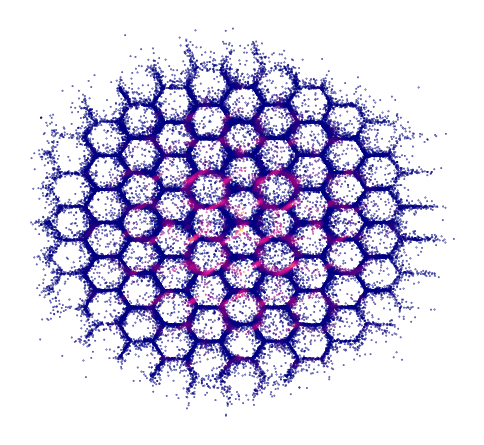} 
        \captionof{subfigure}{EDM}
    \end{minipage}%
    \begin{minipage}{0.25\linewidth}
        \centering
        \includegraphics[width=\linewidth]{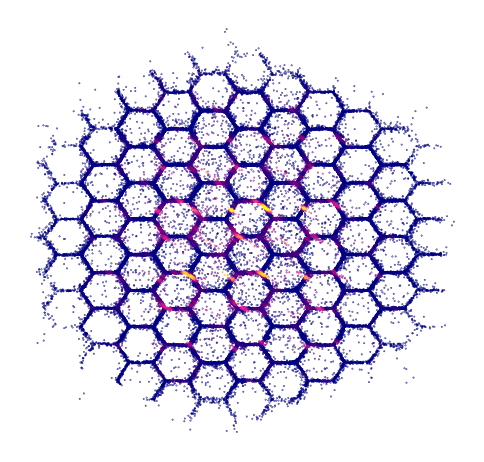} 
        \captionof{subfigure}{LogSNR}
    \end{minipage}%
    \begin{minipage}{0.25\linewidth}
        \centering
        \includegraphics[width=\linewidth]{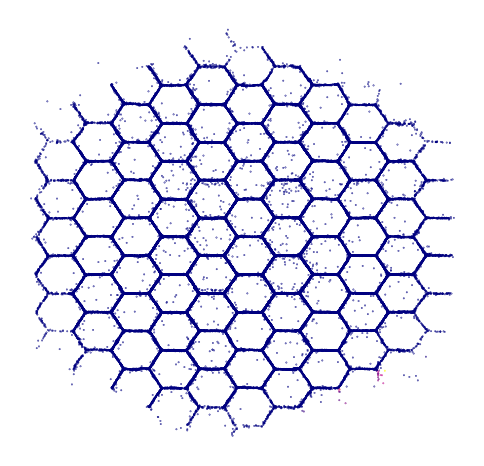} 
        \captionof{subfigure}{AYS}
    \end{minipage}%
    \caption{
        \textbf{Modeling a 2D toy distribution}: (a) Ground truth samples; (b), (c), and (d) are samples generated using 8 steps of SDE-DPM-Solver++(2M) with EDM, LogSNR, and AYS schedules, respectively. Each image consists of 100,000 sampled points. 
    }    
    \label{fig:toy_beehive}
\end{figure}

We further consider two more complex 2D distributions, and use a continuous-time EDM-based diffusion model to learn the score. For these datasets, we train the score model for 100,000 steps with a batch size of 8092. \Cref{fig:toy_beehive,fig:toy_spiral} showcase samples drawn from these models using different schedules.

\rnew{
For these 2D toy data experiments, the colors in \Cref{fig:toy_8x8,fig:toy_8x4,fig:toy_6x6,fig:toy_beehive,fig:toy_spiral} denote the local density of the samples where warmer colors correspond to higher density regions. The density is obtained with a 2D histogram with 50 bins on each axis. All experiments are done in the unconditional generation setting and do not involve any class labels.
}

\begin{table}[h]
\centering
\caption{Performance (measures in negative log likelihood) for mixture of gaussian data and varying solvers, schedules, and number of steps.}
\label{table:toy_nll}
\begin{tabular}{@{}ccccccc@{}}
\toprule
Dataset  & Solver   & Schedule  & NFE=6    & NFE=8    & NFE=10     \\ 
\midrule
\multirow{3}{*}{Gaussian mixture 8x8} & \multirow{3}{*}{SDE-DPM-Solver++(2M)} & EDM      & 9.018    & 4.029   & 1.522   \\
                     &                                       & LogSNR   & 6.250    & 1.834   & 0.071   \\
                     &                                       & AYS      & \textbf{-0.143}   & \textbf{-0.505}  & -\textbf{0.574}   \\
\midrule
\multirow{3}{*}{Gaussian mixture 8x4} & \multirow{3}{*}{DDIM}                 & EDM      & 1.536    & -0.144   & -1.115   \\
                     &                                       & LogSNR   & 1.446    & -0.288   & -1.091   \\
                     &                                       & AYS      & \textbf{-1.999}   & \textbf{-2.260}   & \textbf{-2.222}   \\
\midrule
\multirow{3}{*}{Gaussian mixture 6x6} & \multirow{3}{*}{Stochastic-DDIM}      & EDM      & 1.166    & -0.606   & -0.996   \\
                     &                                       & LogSNR   & -0.012   & -0.978   & -1.231   \\
                     &                                       & AYS      & \textbf{-1.152}   & \textbf{-1.376}   & \textbf{-1.554}   \\
\bottomrule
\end{tabular}
\end{table}

\subsection{CIFAR10, FFHQ, and ImageNet Details} \label{appendix:edm_fid_tables} 
For these experiments, we generate 50,000 images to perform the evaluations. For the continuous-time models, i.e. the CIFAR10 and FFHQ experiments, we use the FID calculation script and reference statistics from \cite{edm}.
For the ImageNet results, we use the evaluation script from \cite{dhariwal2021diffusion}.

We provide more comprehensive results for CIFAR10 and FFHQ in \Cref{table:cifar10_fids,table:ffhq_fids} respectively. As can be seen from the results, the schedules optimized using the KLUB derived for Stochastic-DDIM generalize well to all stochastic solvers. This trend continues to ODE solvers as well, and KLUB-optimized schedules improve results on the first-order DDIM and the multi-step second-order DPM-Solver++(2M) as well. 

\begin{table}[H] 
\caption{Sample fidelity measured by FID $\downarrow$ on the CIFAR10 $32\times 32$ unconditional dataset.}
\label{table:cifar10_fids}
\begin{center}
\begin{tabular}{clccccc}
\hline
                                       & Sampling method                          & Schedule  & NFE=10               & NFE=20               & NFE=30               & NFE=50              \\ \hline
\multirow{10}{*}{Stochastic Sampling}  & \multirow{2}{*}{Stochastic DDIM}         & EDM       & 51.45            & 23.67            & 14.19            & 7.75            \\
                                       &                                          & AYS       & \textbf{33.52}   & \textbf{14.16}   & \textbf{8.78}    & \textbf{5.45}   \\ \cline{2-7}
                                       & \multirow{2}{*}{SDE-DPM-Solver++ (2M)}   & EDM       & 15.32            & 4.64             & 3.15             & 2.64            \\
                                       &                                          & AYS       & \textbf{8.16}    & \textbf{3.23}    & \textbf{2.55}    & \textbf{2.40}   \\ \cline{2-7} 
                                       & \multirow{2}{*}{ER-SDE-Solver 1}         & EDM       & 17.97            & 6.70             & 4.31             & 3.02            \\
                                       &                                          & AYS       & \textbf{12.93}   & \textbf{5.09}    & \textbf{3.50}    & \textbf{2.66}   \\ \cline{2-7} 
                                       & \multirow{2}{*}{ER-SDE-Solver 2}         & EDM       & 9.92             & 3.33             & 2.48             & 2.16            \\
                                       &                                          & AYS       & \textbf{7.77}    & \textbf{3.14}    & \textbf{2.40}    & \textbf{2.14}   \\ \cline{2-7}
                                       & \multirow{2}{*}{ER-SDE-Solver 3}         & EDM       & 9.47             & 3.15             & 2.39             & 2.13            \\
                                       &                                          & AYS       & \textbf{7.55}    & \textbf{3.07}    & \textbf{2.36}    & 2.13            \\ \hline
\multirow{4}{*}{Deterministic Solvers} & \multirow{2}{*}{DDIM}                    & LogSNR    & 16.44            & 6.01             & 3.97             & 2.82            \\
                                       &                                          & AYS       & \textbf{10.73}   & \textbf{4.67}    & \textbf{3.30}    & \textbf{2.56}   \\ \cline{2-7}
                                       & \multirow{2}{*}{DPM-Solver++ (2M)}       & LogSNR    & 5.07             & 2.37             & 2.12             & 2.04            \\
                                       &                                          & AYS       & \textbf{2.98}    & \textbf{2.10}    & \textbf{2.02}    & \textbf{2.01}   \\ 
                                       \hline
\end{tabular}
\end{center}
\end{table}

\begin{table}[H] 
\caption{Sample fidelity measured by FID $\downarrow$ on the FFHQ $64\times 64$ dataset.}
\label{table:ffhq_fids}
\begin{center}
\begin{tabular}{clccccc}
\hline
                                       & Sampling method                          & Schedule  & NFE=10                      & NFE=20                    & NFE=30                    & NFE=50                   \\ \hline
\multirow{10}{*}{Stochastic Sampling}  & \multirow{2}{*}{Stochastic DDIM}         & EDM       & 53.83                   & 31.97                 & 22.14                 & 13.42                \\ 
                                       &                                          & AYS       & \textbf{42.03}          & \textbf{22.73}        & \textbf{14.90}        & \textbf{9.135}       \\ \cline{2-7}
                                       & \multirow{2}{*}{SDE-DPM-Solver++ (2M)}   & EDM       & 23.04                   & 9.67                  & 5.96                  & 3.85                 \\ 
                                       &                                          & AYS       & \textbf{14.79}          & \textbf{5.65}         & \textbf{3.97}         & \textbf{3.13}        \\ \cline{2-7}
                                       & \multirow{2}{*}{ER-SDE-Solver 1}         & EDM       & 21.25                   & 9.29                  & 6.24                  & 4.28                 \\ 
                                       &                                          & AYS       & \textbf{15.27}          & \textbf{7.09}         & \textbf{4.88}         & \textbf{3.68}        \\ \cline{2-7}
                                       & \multirow{2}{*}{ER-SDE-Solver 2}         & EDM       & 12.51                   & 4.49                  & 3.23                  & 2.68                 \\ 
                                       &                                          & AYS       & \textbf{9.04}           & \textbf{4.04}         & \textbf{3.03}         & 2.68                 \\ \cline{2-7}
                                       & \multirow{2}{*}{ER-SDE-Solver 3}         & EDM       & 11.97                   & 4.18                  & 3.06                  & \textbf{2.61}        \\ 
                                       &                                          & AYS       & \textbf{8.71}           & \textbf{3.92}         & \textbf{2.97}         & 2.65                 \\ \hline
\multirow{4}{*}{Deterministic Solvers} & \multirow{2}{*}{DDIM}                    & EDM       & 18.37                   & 8.19                  & 5.60                  & 3.96                 \\
                                       &                                          & AYS       & \textbf{12.83}          & \textbf{6.05}         & \textbf{4.41}         & \textbf{3.38}        \\ \cline{2-7}
                                       & \multirow{2}{*}{DPM-Solver++ (2M)}       & LogSNR    & 7.07                    & 3.41                  & 2.87                  & 2.62                 \\
                                       &                                          & AYS       & \textbf{5.43}           & \textbf{3.29}         & 2.87                  & 2.62                 \\ 
                                       \hline
\end{tabular}
\end{center}
\end{table}

\begin{figure}[h]
    \centering
    \begin{minipage}{0.4\linewidth}
        \centering
        \includegraphics[width=\linewidth]{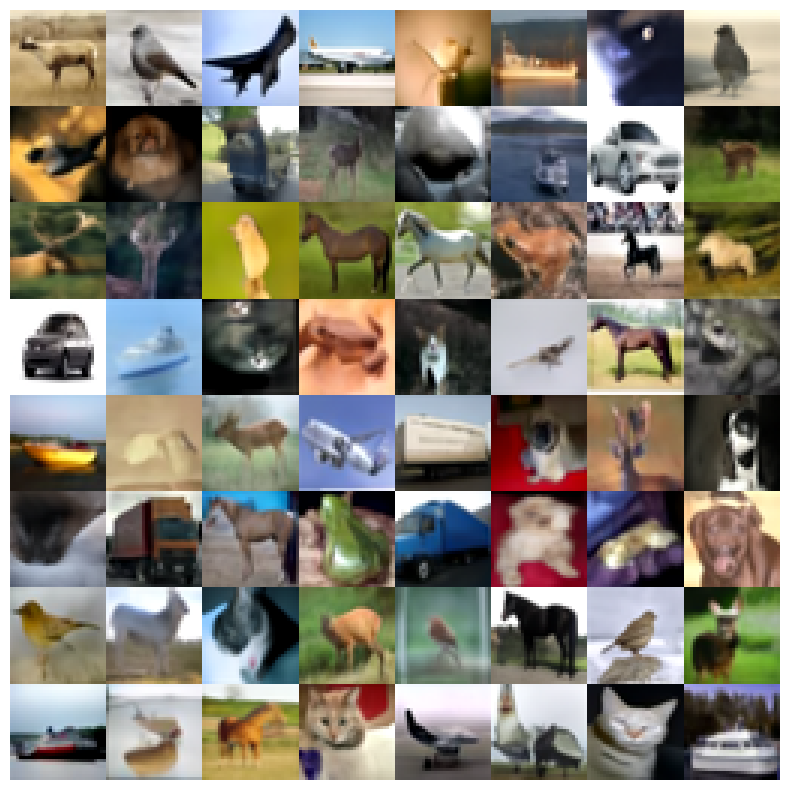}
        \captionof{figure}{EDM Schedule} %
    \end{minipage}%
    \begin{minipage}{0.4\linewidth}
        \centering
        \includegraphics[width=\linewidth]{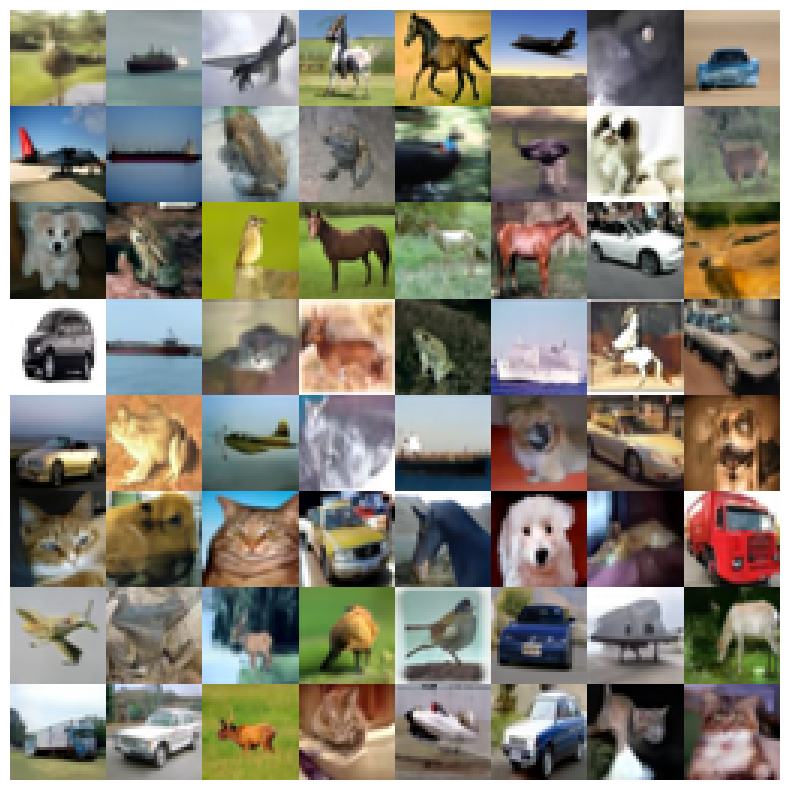}
        \captionof{figure}{AYS Schedule} %
    \end{minipage}
    \caption{Side-by-side comparisons for CIFAR10 with EDM and AYS schedules. Samples are generated using 10 steps with the SDE-DPM-Solver++(2M) solver. }
    \label{fig:cifar10_comparisons}
\end{figure}

\begin{figure}[h]
    \centering
    \begin{minipage}{0.4\linewidth}
        \centering
        \includegraphics[width=\linewidth]{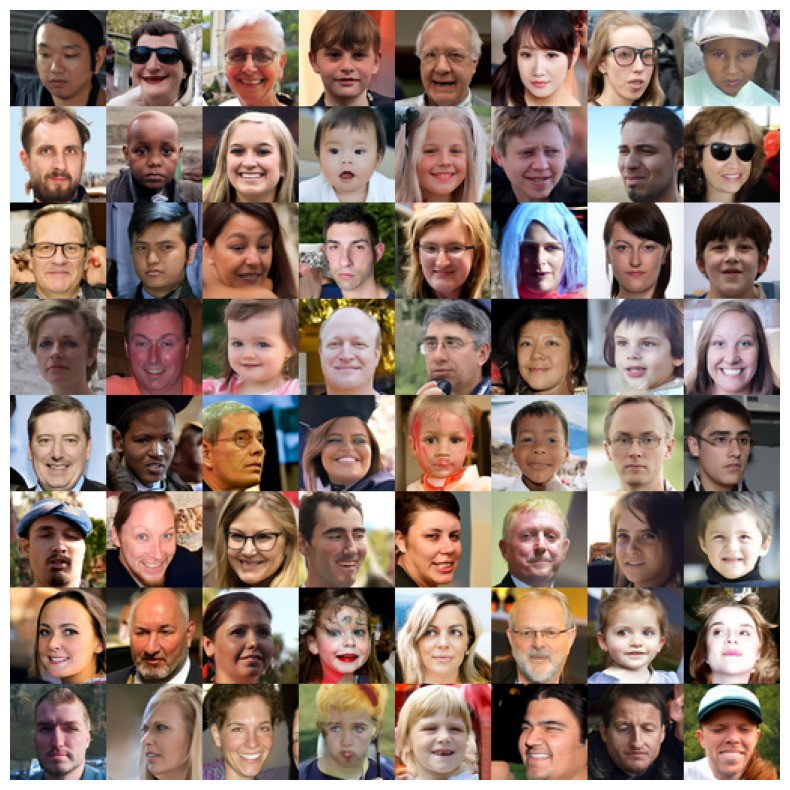}
        \captionof{figure}{EDM Schedule} %
    \end{minipage}%
    \begin{minipage}{0.4\linewidth}
        \centering
        \includegraphics[width=\linewidth]{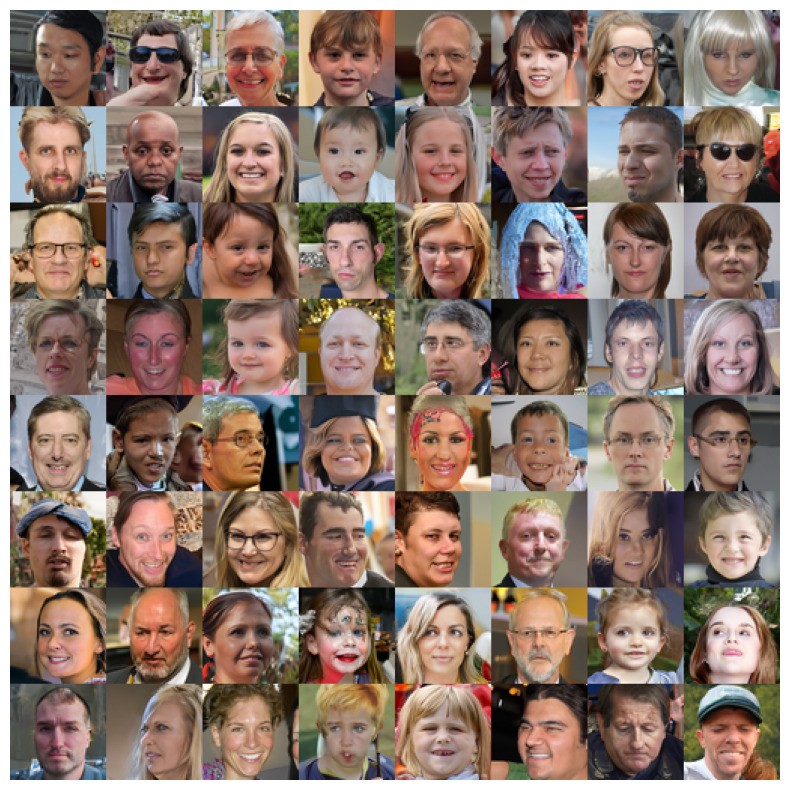}
        \captionof{figure}{AYS Schedule} %
    \end{minipage}
    \caption{Side-by-side comparisons for FFHQ with EDM and AYS schedules.Samples are generated using 10 steps with the DPM-Solver++(2M) solver. }
    \label{fig:ffhq_comparisons}
\end{figure}

\subsection{Comparison with Watson et al.} \label{appendix:google_comparisons}
In this section, we compare our method against the one proposed by \citep{Watson2022LearningFS} on the unconditional ImageNet $64 \times 64$ dataset.
Their approach works by formulating the weights of a multi-step solver and the sampling schedule as trainable parameters, and optimizing them by differentiating through Kernel Inception Distance (KID) as a perceptual loss. Note that this is only applicable to image diffusion models, and cannot be generally used for other data types. Furthermore, it is not clear how their method affects the diversity of samples, due to them directly optimizing the denoising variance to only increase the image quality. 

The authors tested their method against DDIM with standard schedules (time-uniform and time-quadratic). For their experiments, they trained a DDPM following \citep{Nichol2021ImprovedDD} with their $L_{\text{hybrid}}$ objective for 3M steps. In contrast, we use the publicly available checkpoint, which was originally trained for 1.5M steps. For evaluation, the evaluation script from \citep{dhariwal2021diffusion} is used.

\Cref{table:google_fids} summarizes the results. 
The numbers show that, despite using a better diffusion model trained for twice as many steps and optimizing the sampler itself, our optimized schedules alone outperform theirs in extremely low NFE regimes. 
However, as NFE increases, the influence of the schedules diminishes, causing the better diffusion model to gain the upper hand. Nevertheless, when comparing the improvements over the baseline time-uniform schedule, our performance, in terms of FID reduction, is on par with theirs.

\begin{table}[h]
\caption{Image quality measured by FID $\downarrow$ / Inception Score $\uparrow$ on the unconditional ImageNet $64 \times 64$ dataset.} 
\label{table:google_fids}
\begin{center}
\resizebox{\textwidth}{!}{%
\begin{tabular}{@{}cllccccc@{}}
\toprule
Model                             & Sampler                 & Schedule               & NFE=5               & NFE=10               & NFE=15               & NFE=20               & NFE=25              \\ 
\midrule
\multirow{3}{*}{3M steps}         & DDIM                    & Time-uniform           & 135.4 / 5.898   & 40.70 / 12.225   & 28.54 / 13.99    & 24.225 / 14.75   & 22.13 / 15.16   \\
                                  & DDIM                    & Time-quadratic         & 409.1 / 1.380   & 148.6 / 5.533    & 67.65 / 9.842    & 45.60 / 11.99    & 36.11 / 13.225  \\ 
                                  & GGDM +PRED              & Optimized Schedule     & \textbf{55.14 / 12.90}   & \textbf{37.32 / 14.76}    & \textbf{24.69 / 17.225}   & \textbf{20.69 /17.92}     & \textbf{18.40 / 18.12}   \\ 
\midrule
\multirow{2}{*}{1.5M steps}       & DDIM                    & Time-uniform           & 145.01 / 5.45   & 42.51 / 11.25    & 30.32 / 12.89    & 26.60 / 13.57    & 24.77 / 14.00   \\
                                  & DDIM                    & AYS                    & \textbf{50.38 / 11.08}   & \textbf{29.23 / 13.64}    & \textbf{24.21 / 14.24}    & \textbf{22.26 / 14.62}    & \textbf{21.42 / 14.80} \\ 
\bottomrule
\end{tabular}%
}
\end{center}
\end{table}

\subsection{
Text-to-Image Extras
} \label{appendix:extra_txt2img}
For these models that rely heavily on classifier-free guidance, each guidance value changes the models outputs, and can be seen as its own model. As such, it would be ideal to optimize the schedule for each guidance value. 
However, to keep things simple, we opt to only optimize the schedule using a default guidance value, and use the same schedule for all guidance weights in these results.

\begin{figure}[h]
    \centering
    \begin{minipage}{0.5\linewidth}
        \centering
        \includegraphics[width=\linewidth]{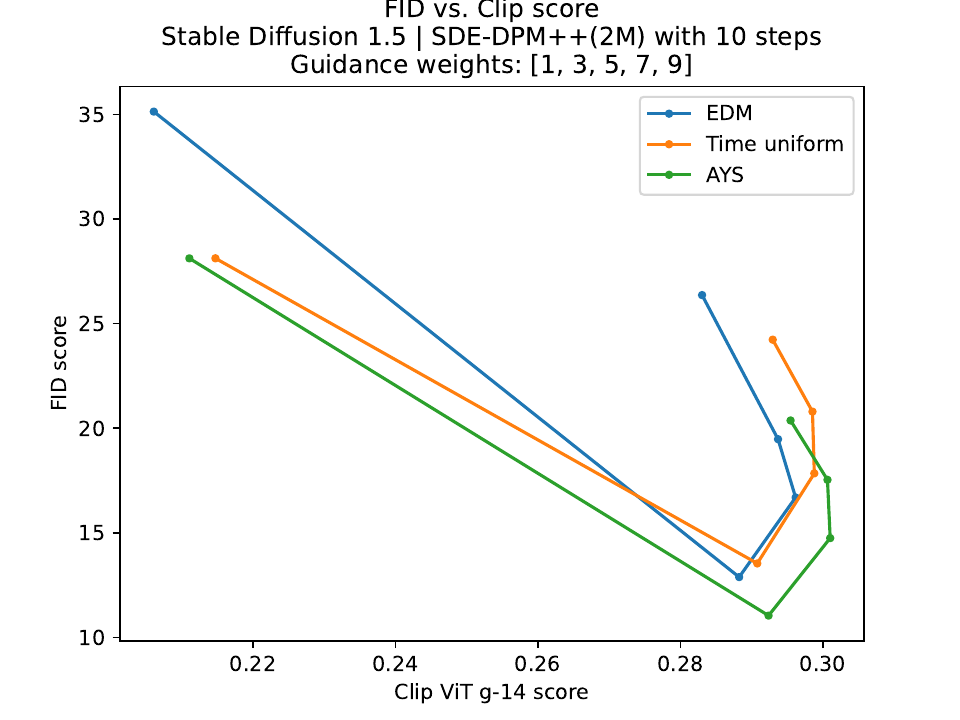}
    \end{minipage}%
    \begin{minipage}{0.5\linewidth}
        \centering
        \includegraphics[width=\linewidth]{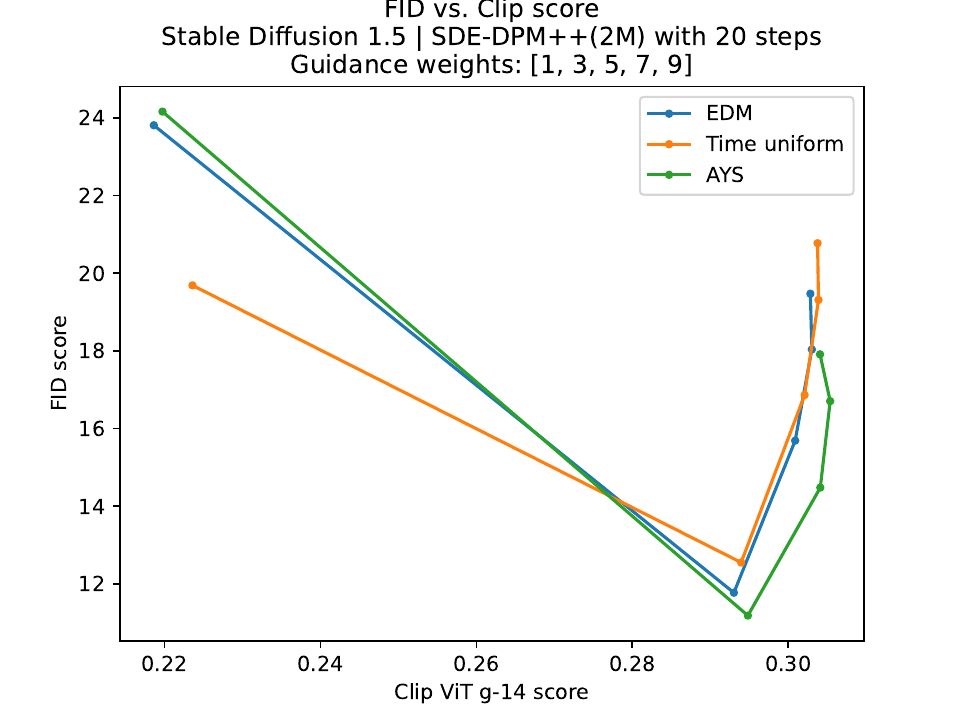}
    \end{minipage}%
    \caption{
        Plotting FID vs. CLIP scores for different classifier-free guidance weights for Stable Diffusion 1.5 using SDE-DPM-Solver++(2M) with 10 and 20 steps. 
    }
    \label{fig:sd15_pareto_curves}
\end{figure}

\begin{figure}[h]
    \centering
    \begin{minipage}{0.5\linewidth}
        \centering
        \includegraphics[width=\linewidth]{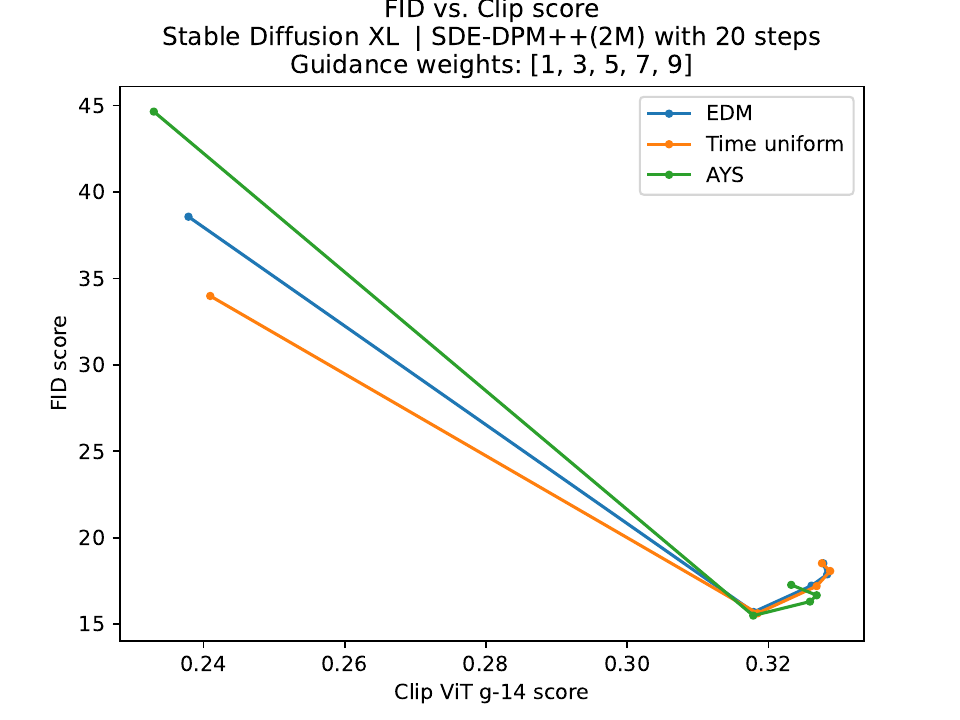} 
    \end{minipage}%
    \begin{minipage}{0.5\linewidth}
        \centering
        \includegraphics[width=\linewidth]{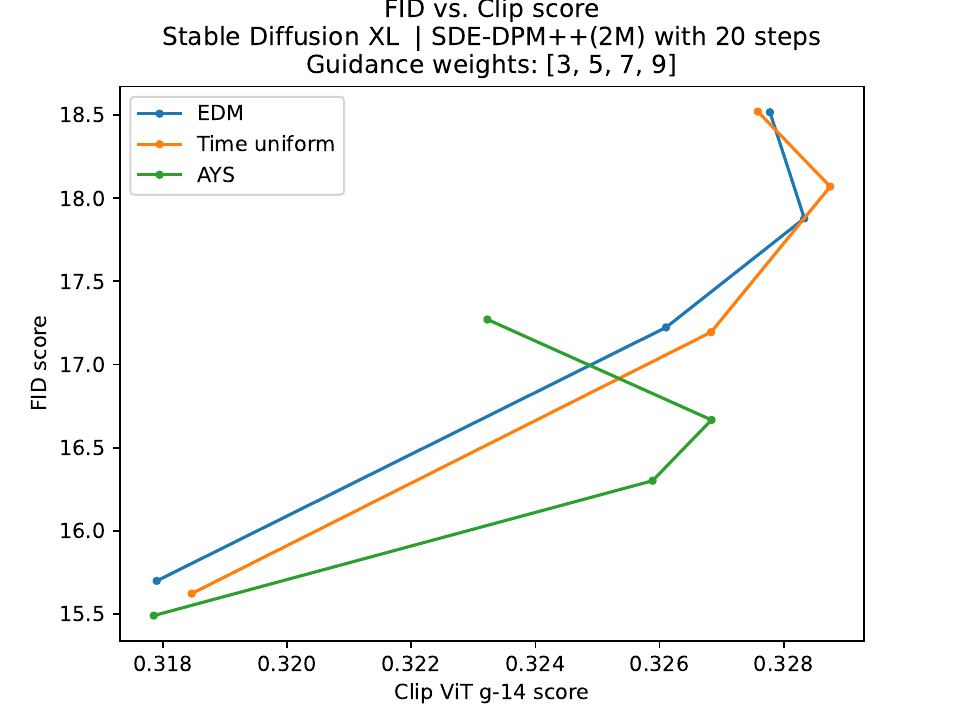} 
    \end{minipage}%
    \caption{
        Plotting FID vs. CLIP scores for different classifier-free guidance weights for SDXL using SDE-DPM-Solver++(2M) with 20 steps. The image on the right is a zoomed in version of the left without the left most point corresponding to no guidance. 
    }
    \label{fig:sdxl_pareto_curves}
\end{figure}

We made use of the COCO~\citep{lin2014microsoft} dataset to optimize the schedule for the text-to-image models. We used a subset of 10,000 images for this task, and excluded these images during FID evaluation. \Cref{fig:sd15_pareto_curves,fig:sdxl_pareto_curves} represent FID vs. CLIP score pareto curves for Stable Diffusion 1.5 and SDXL respectively. 

\begin{figure}[H]
    \centering
    \includegraphics[width=0.8\linewidth]{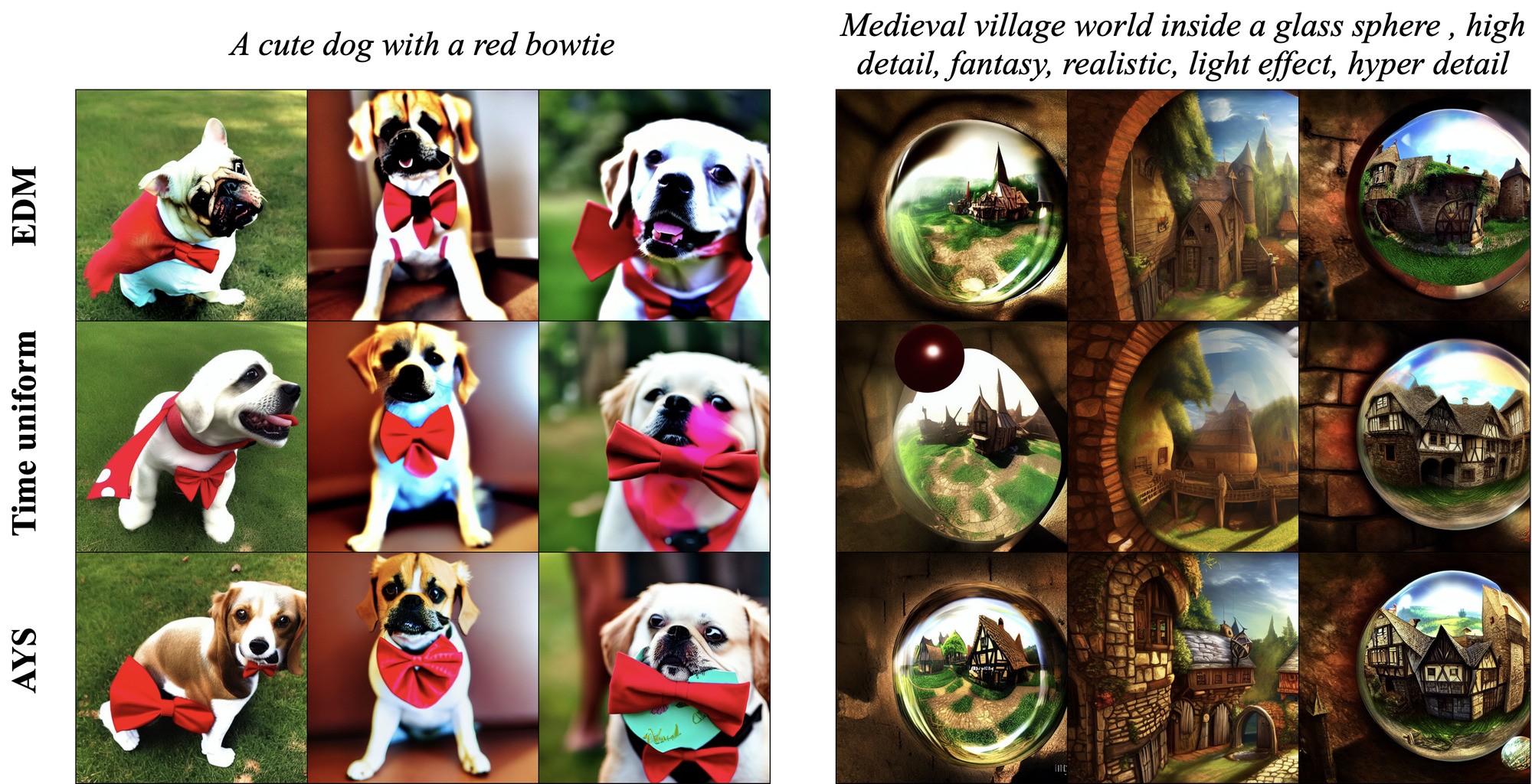}
    \caption{SD 1.5 + 10 steps + SDE-DPM-Solver++(2M)}
    \label{fig:extra_sd15_outputs}
\end{figure}

\begin{figure}[h]
    \centering
    \includegraphics[width=0.8\linewidth]{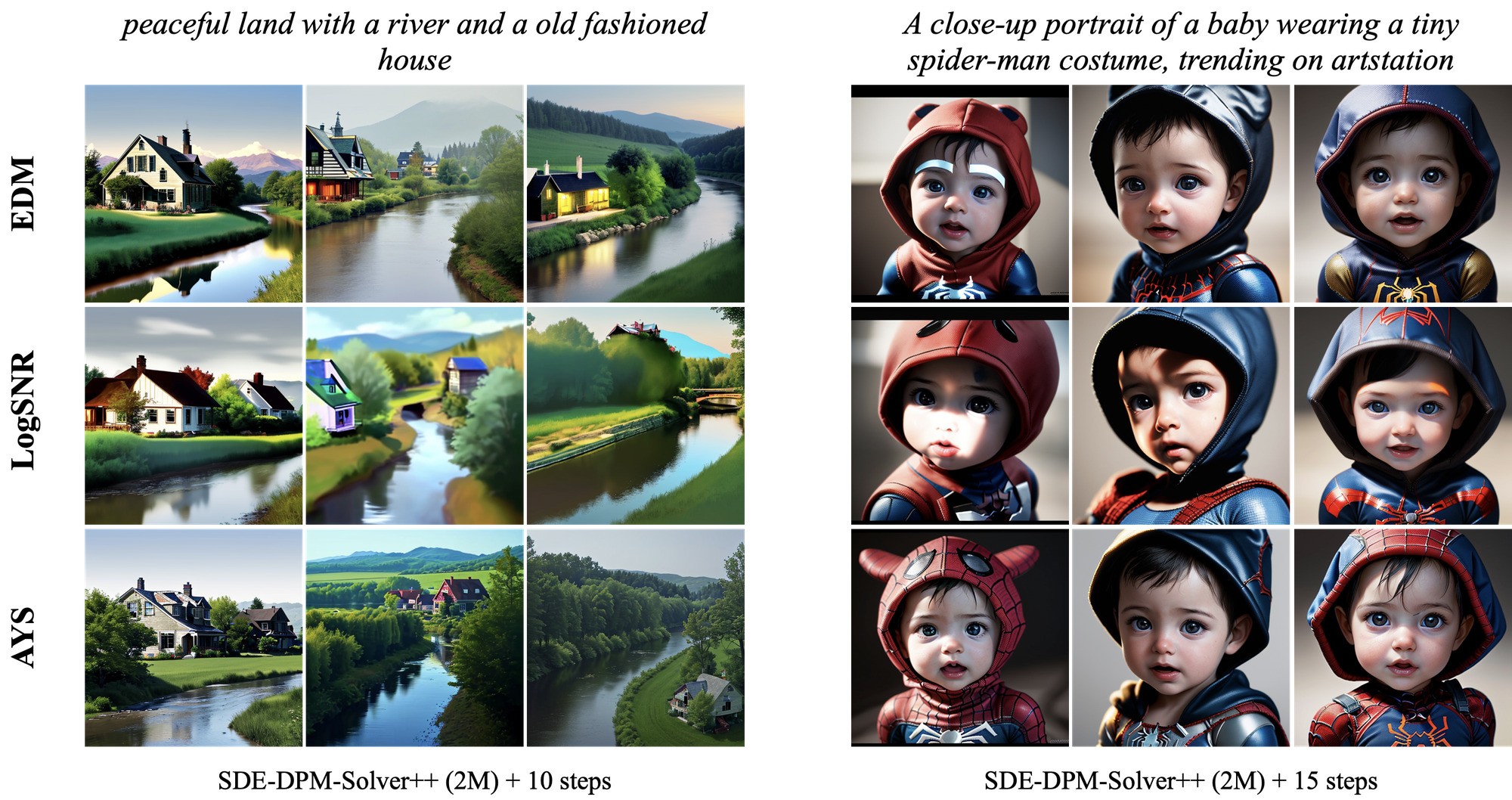}
    \caption{DreamShaper + SDE-DPM-Solver++(2M)}
    \label{fig:dreamshaper_outputs}
\end{figure}

\begin{figure}[h]
    \centering
    \includegraphics[width=0.8\linewidth]{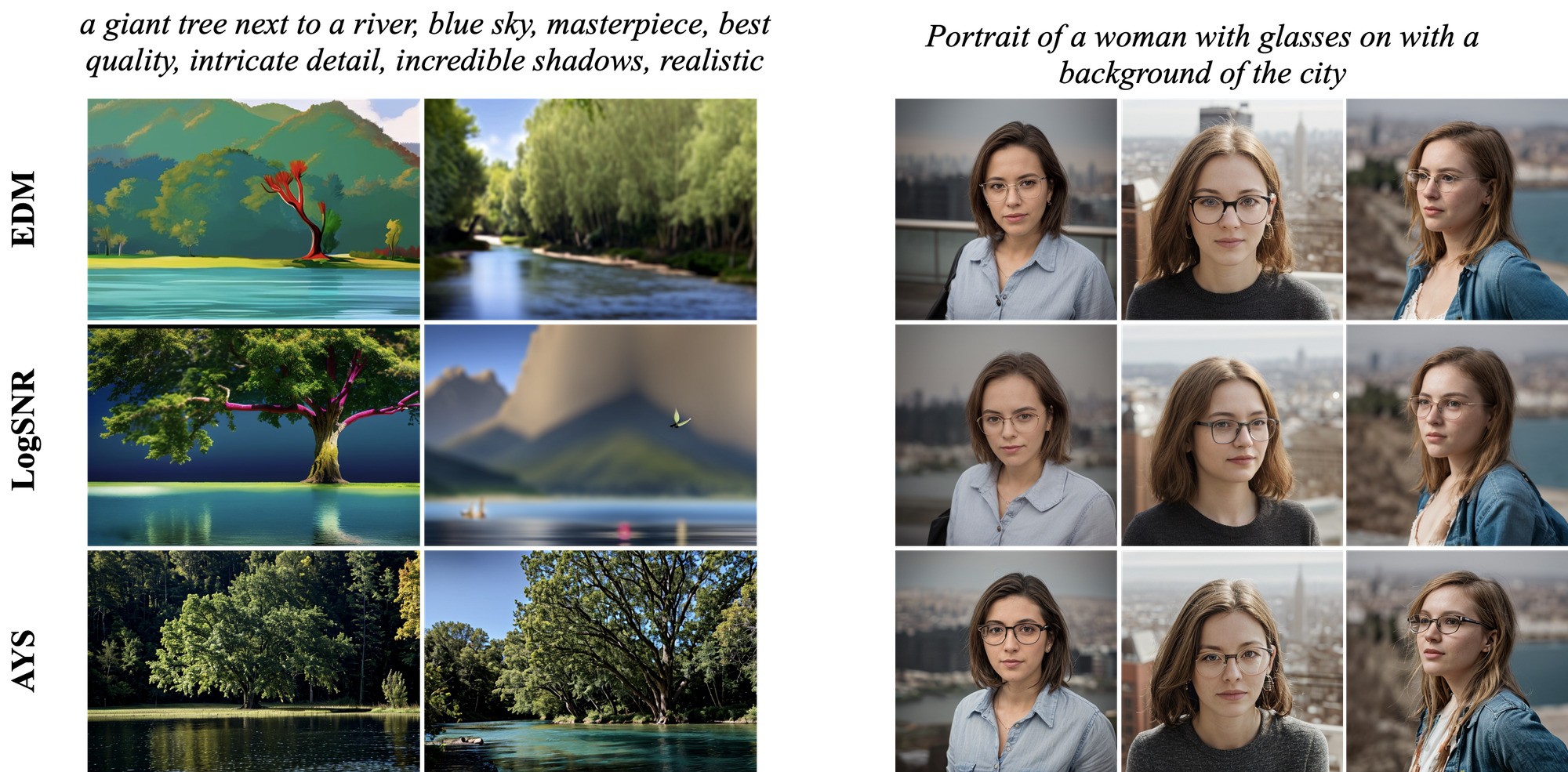}
    \caption{RealisticVision 5.1 + 10 steps + SDE-DPM-Solver++(2M)}
    \label{fig:realisticVision_outputs}
\end{figure}

Interestingly, our optimized SD 1.5 schedule also generalizes and improves images for several personalized text-to-image models based on Stable Diffusion 1.4/1.5. \Cref{fig:extra_sd15_outputs,fig:dreamshaper_outputs,fig:realisticVision_outputs} show some side-by-side comparisons for these models. Please visit our \href{https://research.nvidia.com/labs/toronto-ai/AlignYourSteps/}{
\textcolor{RubineRed}{project page}} for additional qualitative examples. 

To quantitatively evaluate the effectiveness of different schedules we performed a user study. See results in \Cref{fig:user_study}. 
This study involved 42 participants and 600 distinct prompts. For each prompt, three images were generated using EDM, Time Uniform, and AYS schedules using SDE-DPM-Solver++(2M) with 10 steps. Participants were asked to choose the best image in terms of fidelity and text alignment. The results, shown in \Cref{fig:user_study}, reveal a clear preference for the optimized schedule. 

\subsection{Stable Video Diffusion Details} \label{appendix:video_model_details}
For the video diffusion experiments, we used the validation subset of the WebVid10M dataset \citep{webvid} to optimize the schedule. This subset contains 5,000 videos from the internet and we downsampled each to a resolution of $320 \times 576$. 
Given the unclear nature as to how other inputs to the model besides the first frame were obtained during the training of SVD, such as motion bucket id and noise augmentation strength, we simply set them to default values in our experiments. Note that this is extremely sub-optimal, as the model was not trained in this way, however it still produced visible benefits in our experiments.

We also do a user-study on the generated videos. For this, we asked ChatGPT for 150 visually interesting prompts. Afterwards, we used DALLE3 and SDXL to generate 150 images from these prompts. These images will act as the first frames of our generated videos. For each image and schedule, we generated 4 videos, resulting in 1200 videos, 600 using EDM and 600 using the optimized schedule. These were shown to users and asked to identify the best one. Results are summarized in \Cref{table:svd_user_study}.

\begin{figure}[H]
    \centering
    \includegraphics[width=0.8\linewidth]{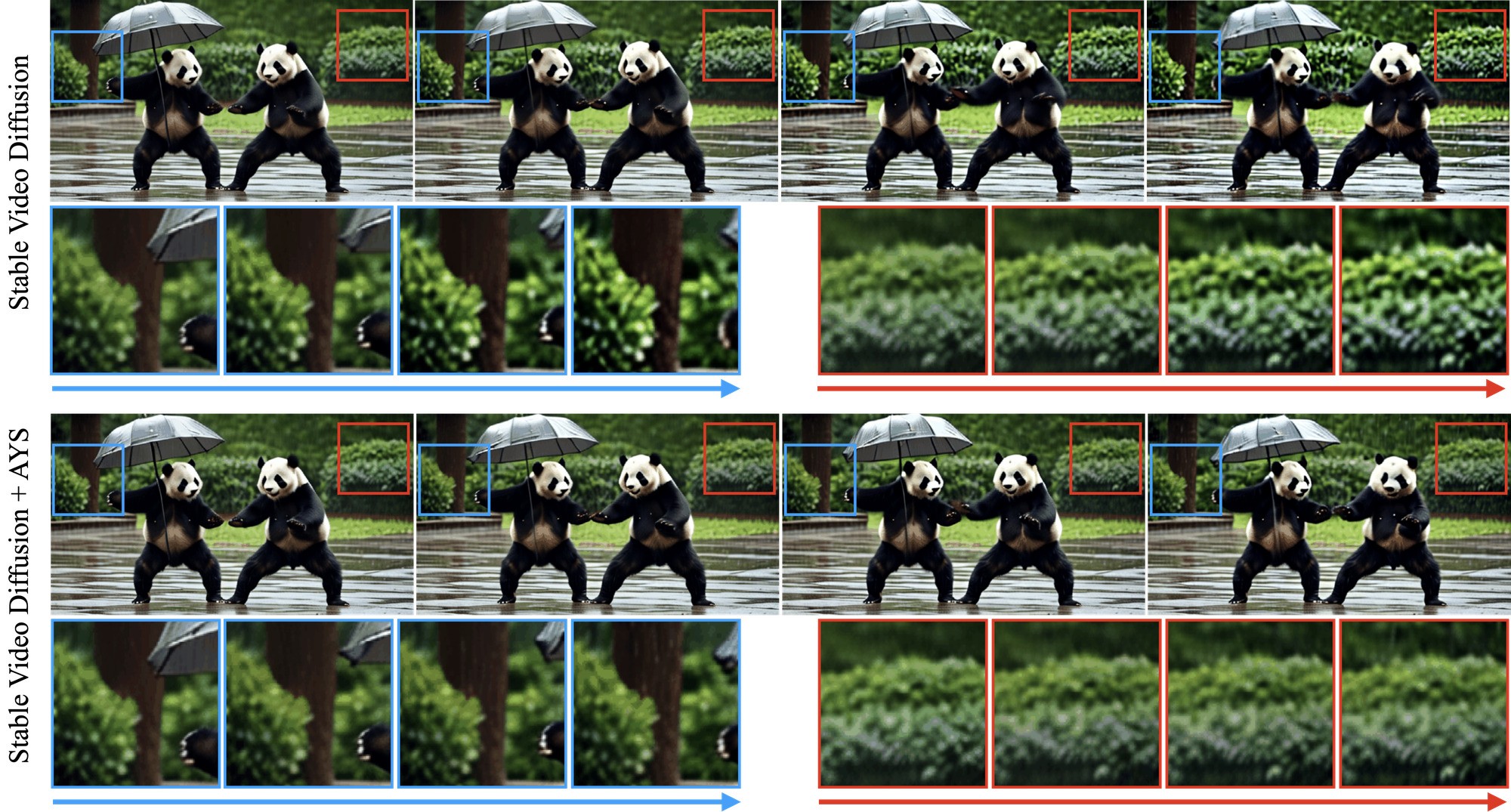}
    \caption{
        Side-by-side comparisons for Stable Video Diffusion \citep{Blattmann2023StableVD}. Using the optimized schedule results in a more stable video; note the temporal color distortions of the background for the baseline. 
    }
    \label{fig:svd_pandas}
\end{figure}

\end{document}